\documentclass{article} 
\usepackage{iclr2025_conference,times}
\usepackage{subcaption}

\usepackage{amsmath,amsfonts,bm}

\def\eqref#1{equation~\ref{#1}}

\def\1{\bm{1}}

\DeclareMathAlphabet{\mathsfit}{\encodingdefault}{\sfdefault}{m}{sl}
\SetMathAlphabet{\mathsfit}{bold}{\encodingdefault}{\sfdefault}{bx}{n}

\usepackage{hyperref}
\usepackage{url}

\usepackage{microtype}
\usepackage{hyperref}
\usepackage{url}
\usepackage{booktabs}

\usepackage{lineno}
\usepackage{amsmath}
\usepackage{amssymb}
\usepackage{mathtools}
\usepackage{wrapfig}
\usepackage{amsthm}
\usepackage{bbm} 
\usepackage{rotating}
\usepackage{algpseudocode}
\usepackage[ruled]{algorithm2e}

\usepackage{float}            

\definecolor{darkblue}{rgb}{0, 0, 0.5}
\hypersetup{colorlinks=true, citecolor=darkblue, linkcolor=darkblue, urlcolor=darkblue}

\title{\textsc{Omni-Thinker}: Scaling Multi-Task RL in LLMs with Hybrid Reward and Task Scheduling}

\author{Derek Li$^{1}$\thanks{Equal contribution. $^{\spadesuit}$Corresponding to: {\{jiaming.zhou, yingxue.zhang\}@huawei.com}}\hspace{2mm}
Jiaming Zhou$^{1\spadesuit}$\footnotemark[1] \hspace{2mm}
Leo Maxime Brunswic$^{1}$\footnotemark[1] \hspace{2mm}
Abbas Ghaddar$^1$\hspace{2mm} Qianyi Sun$^1$ \hspace{2mm} \\
\textbf{Liheng Ma}$^2$\hspace{2mm}
\textbf{Yu Luo}$^3$\hspace{2mm} 
\textbf{Dong Li}$^3$\hspace{2mm} 
\textbf{Mark Coates}$^2$\hspace{2mm}
\textbf{Jianye Hao}$^3$\hspace{2mm} 
\textbf{Yingxue  Zhang}$^{1\spadesuit}$\hspace{2mm}
\\
$^1$ Huawei Noah’s Ark Lab, Montréal, Canada \\
$^2$ McGill University and Mila - Québec AI Institute \\
$^3$ Huawei Noah’s Ark Lab, Beijing, China \\
}

\newtheorem{theorem}{Theorem}

\newtheorem{defi}{Definition}
\newtheorem{assumption}{Assumption}

\iclrfinalcopy
\begin{document}
\maketitle
\begin{abstract}
The pursuit of general-purpose artificial intelligence depends on large language models (LLMs) that can handle both structured reasoning and open-ended generation. We present \textsc{Omni-Thinker}, a unified reinforcement learning (RL) framework that scales LLMs across diverse tasks by combining hybrid rewards with backward-transfer–guided scheduling. Hybrid rewards integrate rule-based verifiable signals with preference-based evaluations from an LLM-as-a-Judge, enabling learning in both deterministic and subjective domains. Our scheduler orders tasks according to accuracy backward transfer (BWT), reducing forgetting and improving multi-task performance. Experiments across four domains show gains of $6.2\%$ over joint training and $12.4\%$ over model merging. Moreover, we demonstrate that simple assumptions on accuracy transfer yield accurate predictions of curriculum outcomes, with entropy dynamics explaining deviations due to generative tasks. These findings underscore the importance of BWT-aware scheduling and hybrid supervision for scaling RL-based post-training toward general-purpose LLMs.
\end{abstract}

\section{Introduction} 

\begin{figure*}[h]
    \centering
    \includegraphics[width=\columnwidth]{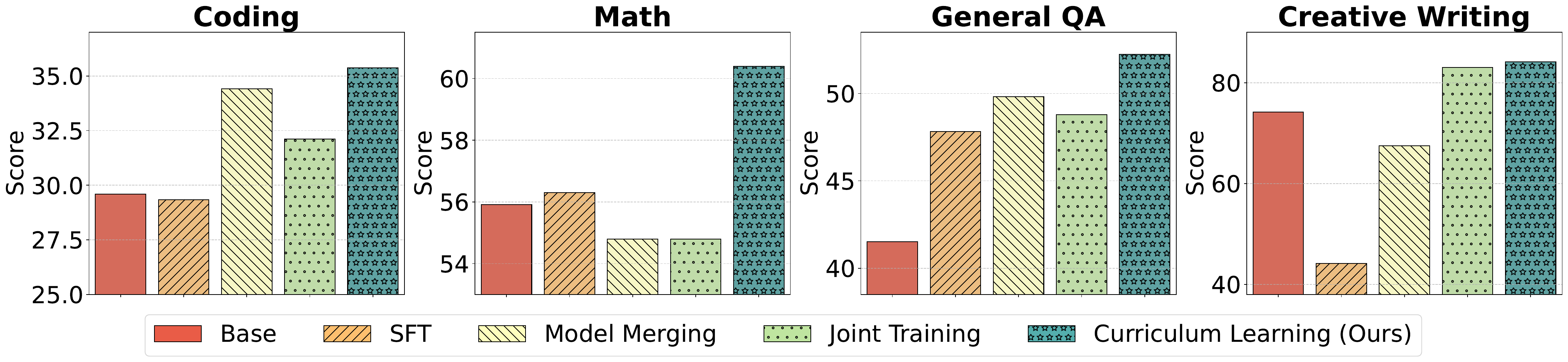}
    \caption{Performance across four task domains, comparing Joint Training and Curriculum Learning against baselines including SFT and Model Merging. Curriculum Learning achieves the strongest results, showing that controlling how tasks are scheduled is crucial for effective multi-task learning.}
    \vspace{-0.2 cm}
    \label{fig:multitask_subplots}
\end{figure*} 

Reinforcement learning (RL) has become an effective approach for improving large language models (LLMs)~\citep{hurst2024gpt,liu2024deepseek,dubey2024llama,yang2024qwen2}, particularly in structured domains such as math and coding where verifiable, rule-based rewards are available~\citep{deepseekai2025deepseekr1incentivizing,deepscaler2025SurpassingO1preview,kimiteam2025kimik15scalingreinforcement}. Methods such as Group Relative Policy Optimization (GRPO)~\citep{shao2024deepseekmath} show that even coarse learning signals can steer LLMs toward structured, chain-of-thought reasoning. However, most RL methods remain tailored to deterministically verifiable tasks, limiting their utility in open-ended domains such as general QA and creative writing. Moreover, training LLMs across multiple tasks remains challenging because it requires optimizing for diverse forms of feedback signals, including binary correctness checks in structured tasks and subjective, preference-based judgments in generative ones.

We address this challenge with \textsc{Omni-Thinker}, a unified RL framework that enables LLMs to learn from both rule-based and generative supervision under a single policy. Building on \textit{Reinforcement Learning with Verified Reward (RLVR)}, our method integrates symbolic verifiers with \textit{LLM-as-a-Judge} evaluations~\citep{zheng2023judging, zhang2024generativerm} to handle subjective tasks. Our curriculum is forgetting-aware; it is guided by backward transfer (BWT), where BWT denotes test-performance backward transfer computed on a normalized, task-specific test metric. Ordering task training according to this signal yields effective curricula across heterogeneous domains. 
We show that the final accuracy of model after curriculum learning is well predicted by {\it forgettability} ranking, even under simplifying assumptions. Empirically, we observe complementary entropy dynamics, fine-tuning on creative writing tends to increase the model's output entropy, whereas training on verifier-supervised, structured tasks tends to decrease it; this trend is consistent with our BWT-guided choice to train structured tasks before open-ended ones.
Across four domains, \textsc{Omni-Thinker} improves generalization while reducing forgetting, with average gains of 6.2\% over joint multi-task training and 12.4\% over model merging, respectively.

Our key contributions are threefold.  (1) We present \textsc{Omni-Thinker}, a unified framework that trains a single policy across four diverse domains, using hybrid verifiable and preference-based rewards.  (2) We develop a forgetting-aware curriculum based on backward transfer (BWT) linear ordering maximization over task-specific test performance to reduce forgetting, outperforming joint multi-task training and model merging. (3) We empirically analyze training dynamics through the lens of entropy, revealing that structured domains (math, coding) systematically decrease output entropy while open-ended domains (creative writing) increase it, thereby providing an explanatory link between entropy evolution and the effectiveness of BWT-guided curricula.

\section{Framework Overview}

We introduce \textsc{Omni-Thinker} as a unified reinforcement learning framework for large language models that integrates hybrid rewards with task scheduling guided by backward transfer. Unlike prior approaches that separate reasoning and generative domains, \textsc{Omni-Thinker} maintains a single policy across heterogeneous tasks, including Math, Coding, General QA, and Creative Writing, while dynamically ordering training to minimize forgetting. The framework is instantiated using Multi-Task GRPO, augmented with both symbolic verifiers and LLM-as-a-Judge supervision, and a curriculum determined by accuracy- and entropy-based backward transfers.

\subsection{Notation and Training Objective}
 We give ourselves a vocabulary $\mathcal V$ with a special end-of-sequence token $\mathrm{eos}$. The set of finite sequences of tokens is denoted $\mathcal V^*$; for any sequence $o\in \mathcal V^*$, its length is denoted $|o|$ and we say that $o$ is {\it complete} if $o_{|o|} = \mathrm{eos}$.
A model, parameterized by $\theta$, defines a conditional distribution $\pi_\theta(y_{t} \mid y_{<t})$ for any given sequence of tokens $(y_t)_{t\in \mathbb N}$. It induces a policy $\pi_\theta^\otimes$ on token sequences defined by $\pi_\theta^\otimes(o \mid q,o_{<t_0}):= \prod_{t=t_0}^{|o|} \pi_\theta(o_t\mid q, o_{<t})$. 
We adopt a multi-task RL (MTRL) formulation: a task is a couple $T=(\mathcal D,R)$ where $\mathcal D$ is a dataset of prompts and $R(q,o)$ is a task-specific reward function.
   Given a set of $K$ tasks $\mathcal{T} = \{T_1, \ldots, T_K\}$, 
  the goal is to learn a unified policy $\pi_\theta$ that maximizes the expected reward over the task distribution:
\begin{equation}
\max_\theta \; \mathcal{J}(\theta) = \mathbb{E}_{(\mathcal D,R) \sim P(\mathcal{T})} \left[\mathbb{E}_{q \sim \mathcal{D}, o \sim \pi_\theta^\otimes(\cdot \mid q)} \left[ R(o) \right]\right],
\label{rl_obj}
\end{equation}
where $P(\mathcal{T})$ is a task sampling distribution, which determines task exposure during training.

In order to train $\pi_\theta$ to maximize the objective $\mathcal J$, we extend the GRPO \citep{deepseekai2025deepseekr1incentivizing} algorithm to the multi-task setting by jointly optimizing over task-specific reward signals and reference policies. For each input prompt $q$, GRPO samples a group of outputs $\{o_{q,1}, o_{q,2}, \cdots, o_{q,G}\}$ from the old policy  $\pi_{\theta_{old}}$. 
A task-specific reward function $R_k(q,o)$ scores each output. The policy $\pi_{\theta}$  is updated to maximize expected return while controlling divergence from a reference policy.

We define the policy ratio $\rho_{q,i,t}$ and the normalized advantage estimate $\hat{A}_{q,i,t}$ as follows:
\begin{eqnarray}
\mu_q = \text{mean}\big(\{ R_k(q,o_{q,i}) \}_{i=1}^G\big),
&&
\sigma_q = \text{std}\big(\{ R_k(q,o_{q,i}) \}_{i=1}^G\big), \\
\rho_{q,i,t} = \frac{\pi_{\theta}(o_{q,i,t} \mid q, o_{q,i,<t})}{\pi_{\theta_{\text{old}}}(o_{q,i,t} \mid q, o_{q,i,<t})}, &&
\hat{A}_{q,i,t} = \frac{R_k(q,o_{q,i}) - \mu_q}{\sigma_q}.
\end{eqnarray}

This allows us to write the MT-GRPO objective as
\begin{equation}
\begin{split}
    &\mathcal{J}_{\text{MT-GRPO}}(\theta) =\mathbb{E}_{k \sim K, q \sim \mathcal{D}_k, \{o_{q,i}\}_{i=1}^G \sim \pi_{\theta_\text{old}}^\otimes(\cdot \mid q_k)} \\ 
    &\frac{1}{G} \sum_{i=1}^G \frac{1}{|o_{q,i}|} \sum_{t=1}^{|o_{q,i}|} \left\{ \min \left[ \rho_{q,i,t} \hat{A}_{q,i}, 
    \text{clip}\left( \rho_{q,i,t}, 1{-}\epsilon, 1{+}\epsilon \right) \hat{A}_{q,i} \right]  - \beta_k \mathbb{D}_{KL}\left[\pi_{\theta} || \pi_{ref}\right]\right\} ,
\end{split}
\label{eq:GRPO-obj}
\end{equation}
where 
\begin{equation}
    \mathbb{D}_{KL}\left[\pi_{\theta} || \pi_{ref}\right] = \frac{\pi_{ref}(o_{q,i,t}|q,o_{q,i,<t})}{\pi_{\theta}(o_{q,i,t}|q,o_{q,i,<t})}- \log\frac{\pi_{ref}(o_{q,i,t}|q,o_{q,i,<t})}{\pi_{\theta}(o_{q,i,t}|q,o_{q,i,<t})} - 1.
\end{equation}

The clipping parameter $\epsilon$ stabilizes updates by keeping policy ratios within a bounded range, following the PPO approach \citep{schulman2017proximal}. The KL divergence term regularizes the new policy towards the reference policy $\pi_{\text{ref}}$, weighted by a task-specific coefficient $\beta_k$.

\subsection{Hybrid Rewards}
We design a hybrid reward system that unifies reinforcement learning across both structured reasoning tasks and open-ended generative domains. 

\paragraph{Verifiable Supervision.} For tasks with objective correctness signals, such as symbolic math and code generation, we define binary rewards based on symbolic matches, test case results, or other deterministic evaluators depending on the tasks.

\paragraph{Short-Form Open-Ended Supervision.}
For language tasks with known or extractable ground-truth answers such as general question answering (QA), we reformulate queries into open-ended prompts and incorporate distractor responses (LLM-generated plausible but incorrect answers) into the context. Instead of labeling options, we prompt the model to reason using the \texttt{<think>}...\texttt{</think>} format and to output answers within \texttt{<answer>}...\texttt{</answer>} tags. Responses are evaluated with a binary reward based on string matching or set membership against reference answers, thereby encouraging semantic grounding and mitigating shallow pattern memorization. We find that conditioning the LLM on a diverse set of candidate options, including one correct answer and multiple distractors, is key to steadily improving general-domain reasoning while reducing susceptibility to random guessing or reward hacking, compared to directly prompting the model to generate open-ended answers during training without the augmented context.

\paragraph{Long-Form Open-Ended Supervision.} For subjective tasks lacking ground truth (e.g., dialogue, writing), we use an \textit{LLM-as-a-Judge} ~\citep{chen2025judgelrm} to assign a scalar reward based on rubric-aligned pairwise preferences between candidate outputs. This enables learning in domains where symbolic correctness is insufficient or intractable. This prompt-based approach leverages recent advances in the general reasoning capabilities of LLMs, using generated chain-of-thoughts to elicit a ternary reward signal, preferred, tie, or dispreferred, without requiring large-scale preference data collection and reward model training.

Together, these components form a unified hybrid reward scheme: verifiable rewards ensure correctness where possible and generative-based signals cover subjective domains. This design enables reinforcement learning to scale across diverse tasks, from reasoning to open-ended generation.

\subsection{Joint training and Curriculum Learning}
\label{sec:methodology_CL}
In practice, a maximization step of the training objective $\mathcal{J}_{\text{MT-GRPO}}$ requires a batch $B$ of prompts sampled from $\bigcup _{k=1}^K \mathcal D_k$ then sampling a batch of outputs $\{o_{q,i}\}_{i=1}^G$ for each $q\in B$.  A multi-task schedule is defined as a sequence of batches  $(B_s)_{s=1}^{s_{\max}}$ 
such that $\forall s\neq s', B_s\cap B_{s'}=\emptyset$ and $\bigcup_{s=1}^{s_{\mathrm{max}}} B_s = \bigcup _{k=1}^K \mathcal D_k$.

Two special cases are considered: {\it Joint Training} and {\it Curriculum Learning}. 
{\it Joint Training} consists in sampling each batch $B$ uniformly at random among all samples (without replacement), disregarding their corresponding tasks: $\forall s, B_s \sim \mathcal U\left(\bigcup_{k=1}^K \mathcal D_k \setminus \bigcup_{s'<s}B_{s'}\right)$.
{\it Curriculum Learning} on the other hand consists of pure batches chosen from the same task until exhaustion of the task dataset. By pure, we mean that each batch is derived from only one task dataset:
$\forall s, B_s \sim \mathcal U\left(\mathcal D_{k_s} \setminus \bigcup_{s'<s}B_{s'}\right)$ for some task schedule $(k_s)_{s\in \{1,\cdots,s_{\mathrm{max}}\}}$.
A Curriculum is described by a permutation $\sigma \in \mathfrak{S}_K$ of the tasks, with $\mathfrak{S}_K$ the set of permutation of $\{1,\cdots,K\}$.

\section{Backward Transfer for Task-scheduling}
We intend to use Backward Transfers (BWT) to guide our choice of curriculum. Following \cite{lopez2017gradient} it is defined as follows.
\begin{defi}[Backward Transfer Matrix] \label{defi:BWT} Let $\theta_0$ be a set of  initial parameters of a model $\pi_\theta$ and let $\theta(\theta_0,T)$ 
be the set of parameters obtained after training $\pi_\theta$ on task $T$ starting from $\theta_0$.
Write $\mathrm{Acc}(\theta,T)$ the accuracy of model $\pi_\theta$ on task $T$.
The backward transfer matrix  is defined by
    \begin{equation}\label{equ:BWT}
    \mathrm{BWT}_{ij}(\theta_0):= \log\mathrm{Acc}(
    {\theta(\theta_0,T_{j})},T_i)-\log \mathrm{Acc} (
    {\theta_0},T_i).
\end{equation}
\end{defi}

\begin{wrapfigure}{r}{0.5\linewidth}
    \centering
        \vspace{-0.4cm}
        \begin{algorithm}[H]
        \caption{Final Accuracy under Assumptions~\ref{ass:constantBWT} and~\ref{ass:saturatingAcc}}
        \label{alg:constantBWT}
        \KwIn{$\mathrm{BWT} \in \mathbb{R}^{K \times K}$. 
        $a_{\text{init}}:=\left(\mathrm{Acc}(\theta_0,T_k)\right)_{k=1}^K \in \mathbb{R}^K$.  
        Curriculum $\sigma \in\mathfrak{S}_K$}
        $a \gets a_{\text{init}}$\;
        \For{$j=1$ {\bf to} $K$}{
            $k \gets \sigma(j)$\;
            $a_k \gets a_{\text{init},k}$\;
            \For{$i=1$ {\bf to} $K$}{
                $a_i \gets a_i \times \exp(\mathrm{BWT}_{ik})$\; 
            }
        } 
        \KwRet $a$\;
        \end{algorithm}
        \vspace{-1cm}
\end{wrapfigure}

\subsection{A priori prediction of terminal accuracies under constant BWT}~
Our goal is to \emph{choose a curriculum order $\sigma$ a priori} by predicting the terminal per–task accuracies \emph{without} training all permutations. We propose a simple predictive model in which (i) inter–task backward transfers are treated as constant in log–accuracy, and (ii) training on the full dataset of a task saturates its self–accuracy. Under these assumptions, terminal accuracies for any order $\sigma$ become computable from quantities measured once at initialization.

\paragraph{Setup.}
Using notations from Section~\ref{sec:methodology_CL}, let $\theta_0$ denote the parameters of the pre–trained model $\pi_\theta$, and let $\theta_s$ be the parameters after $s$ optimization steps following a curriculum order $\sigma \in \mathfrak{S}_K$.

\begin{assumption}[Constant off–diagonal BWT in log–accuracy]\label{ass:constantBWT}
For all $i\neq j$ and all optimization steps $s$ along the schedule,
\begin{equation}
\mathrm{BWT}_{ij}\!\left(\theta_{s}\right)
= \mathrm{BWT}_{ij}\!\left(\theta_0\right).
\end{equation}
\end{assumption}

\begin{assumption}[Task–wise saturation]\label{ass:saturatingAcc}
Let $\{B_s\}$ be the sequence of mini–batches processed along the schedule. If, between steps $s_1$ and $s_2$, the full dataset $\mathcal D_k$ of task $T_k$ has been seen, then accuracy saturates on task $T_k$ to the same accuracy as training from $\theta_0$:
\begin{equation}
\bigcup_{s=s_1}^{s_2} B_s = \mathcal D_k
\quad\Rightarrow\quad
\mathrm{Acc}\!\left(\theta_{s_2},T_k\right)
= \mathrm{Acc}\!\big(\theta(\theta_{0},T_k),\,T_k\big).
\end{equation}

\end{assumption}

\begin{theorem}\label{thm:constantBWT_predict}
Under Assumptions~\ref{ass:constantBWT} and~\ref{ass:saturatingAcc}, for any curriculum order $\sigma \in \mathfrak{S}_K$ starting from $\theta_{0}$, the terminal accuracies $\{\mathrm{Acc}(\theta_{s_{\max}},T_k)\}_{k=1}^K$ given the initialization accuracies $\{\mathrm{Acc}(\theta_{0},T_k)\}_{k=1}^K$, the $\mathrm{BWT}(\theta_{0})$ and a curriculum $\sigma$ is exactly the output of Algorithm~\ref{alg:constantBWT}  
\end{theorem}

\paragraph{Reasonableness and limitations.}
Assumption~\ref{ass:constantBWT}  abstracts away known drivers of transfer, such as domain overlap, stochasticity, and entropy evolution, thus is a \emph{toy} yet useful approximation for \emph{a priori} curriculum selection. See Sections~\ref{sec:entropy_discussion} and~\ref{sec:entropy_support} for a discussion of a correction coming from entropy. Working in log–accuracy space keeps accuracies positive but does not eliminate the risk of unrealistic growth over long curricula (e.g., predictions exceeding $1$ in accuracy). Assumption~\ref{ass:saturatingAcc} is reasonable when each task dataset is sufficiently large and optimization keeps the model in a well–conditioned regime, conditions we satisfy in our experiments. Together, these assumptions yield a tractable predictor that captures coarse curriculum effects while remaining simple enough to evaluate without exhaustive training.

\subsection{Curriculum Choice via Linear Ordering Maximization}

Algorithm~\ref{alg:constantBWT} admits the following closed form for the predicted terminal accuracies:
\begin{equation}
\log \mathrm{Acc}(\sigma) - \log \mathrm{Acc}(\mathrm{Id})
= \sum_{i<j} \bigl(\Phi_{\sigma}^{-1}\,\mathrm{BWT}\,\Phi_{\sigma}\bigr)_{ij},
\quad\text{with}\quad 
\Phi_{\sigma,ij} = \mathbf{1}_{i=\sigma(j)}.
\end{equation}

\begin{wrapfigure}{r}{0.5\linewidth}
    \centering
    \vspace{-.5cm}
    \begin{algorithm}[H]
    \caption{Greedy BWT-LOM Curriculum}
    \label{alg:lom}
    \KwIn{BWT matrix $\mathrm{BWT} \in \mathbb{R}^{K\times K}$}
    $\sigma \gets$ empty list\;
    \While{there are unvisited tasks}{
        $k^* \gets \arg\max_{k \notin \sigma}\ \sum_{i\notin \sigma\cup\{k\}} \mathrm{BWT}_{ik}$\;
        append $k^*$ to $\sigma$\;
    }
    \KwRet{$\sigma$}\;
    \end{algorithm}
    \vspace{-.62cm}
\end{wrapfigure}
In words, curriculum reordering acts by permuting the BWT matrix with $\Phi_\sigma$, and the gain relative to the identity schedule is simply the sum of the upper--triangular entries of the permuted matrix.

Given an aggregated score of the form
\begin{equation}
\mathcal S := \sum_{T\in \mathcal T} \alpha_T \log \mathrm{Acc}(\theta,T),
\end{equation}
identifying the best task order amounts to solving a \emph{Linear Ordering Problem} (LOP). see \citep{floudas2008encyclopedia} for an overview. This problem is known to be NP--hard, but for small numbers of tasks ($K$) it can be solved exactly. For larger $K$, a wide range of approximation algorithms and heuristics exist.
A simple heuristic is to rank tasks by a \emph{forgettability score}:
$
F_k := \alpha_k \sum_{i\neq k} \mathrm{BWT}_{ik}
$.
Intuitively, ordering tasks by decreasing $F_k$ prioritizes those that exert the least destructive interference on others (or even provide positive transfer), thereby reducing overall forgetting. 
Our curriculum orders tasks by decreasing column mean of BWT.

\section{Experimental Setup}
\paragraph{Training Datasets.} We curate a multi-domain training dataset covering Math, Coding, General QA, and Creative Writing, with each domain selected to support hybrid reward functions and robust evaluation. For \textbf{Math}, we begin with the OpenR1-Math \citep{openr1} dataset, retaining only word problems and excluding questions that require visual reasoning. We further subsample 12,000 examples to fit our compute budget. For \textbf{Coding}, data is sourced from the code-r1-12k \citep{code-r1} dataset, with outliers exceeding 1024 tokens removed. Each entry includes a code prompt and JSON-formatted unit tests for automatic validation. For \textbf{General QA}, we subsample 5,500 queries from from SuperGPQA \citep{pteam2025supergpqascalingllmevaluation} dataset, proportionally by question category. Each sample comprises a factual question paired with a plain-text answer. We then generate 15 additional confusion options while making sure the uniqueness of correctness by prompting an LLM. 
The \textbf{Creative Writing.} domain leverages 6,650 conversations from Nitral AI’s ShareGPT dataset~\citep{nitral2025creative}, focused on single-turn completions. Samples exceeding two dialogue turns are filtered out, and responses are judged via an \textit{LLM-as-a-Judge} framework.

\paragraph{Evaluation.} We assess performance in each domain using dedicated benchmarks aligned with the task’s evaluation criteria.
\textbf{Math}: accuracy on AIME24~\citep{MAA2024_AIME}, AMC23~\citep{MAA2023_AMC}, Gaokao2023EN~\citep{liao2024mario}, MATH-500~\citep{hendrycksmath2021}, MinervaMath~\citep{lewkowycz2022solving}, and OlympiadBench~\citep{he2024olympiadbench}. 
\textbf{Coding}: pass@1 on BigCodeBench (Complete-Full)~\citep{zhuo2024bigcodebench} and LiveCodeBench (24Oct–25Jan)~\citep{jain2024livecodebenchholisticcontaminationfree}. 
\textbf{General QA}: exact-match accuracy on MMLU-Pro~\citep{wang2024mmlupro}. 
\textbf{Creative Writing}: win rate on the \emph{role-play} and \emph{creative writing} subsets of MT-Bench~\citep{zheng2023judging}, against GPT-4 (\textit{pre-gen, June 16, 2023}).

\paragraph{Baselines.} We use Qwen2.5-7B-Instruct~\citep{yang2024qwen2} as the base model for all experiments, owing to its strong instruction-following capability, which makes it well-suited for reinforcement learning on both structured reasoning and open-domain QA tasks. 
\textbf{Supervised Fine-Tuning (SFT):} In order to have a meaningful comparison with GRPO, we adopt a similar self-sampled data curation and fine-tuning approach with Rejection sampling Fine-Tuning \citep{yuan2023scaling}. We first prompt the base model to generate 128 chain-of-thought responses for our training dataset to ensure we end up with at least one correct response for most queries, then filter them based on the same accuracy reward signals used in GRPO training. We then perform sft on base model using these self-distilled responses. This provides a strong on-policy learning baseline that incorporates explicit reasoning steps through self-distillation from the base model.
\textbf{Model Merging:} We employ \textit{TIES-Merging} ~\citep{yadav2023ties} as our model‑merging baseline. It is a simple yet effective method designed specifically for the multi-task setting that takes into consideration the interference between parameters from models trained on individual tasks during the merging process. It has demonstrated superior performance in multi-task learning compared to linear and task arithmetic approaches \citep{yadav2023tiesmergingresolvinginterferencemerging}. To begin with, we conduct single-task GRPO training using individual task datasets and collect the model weights of the best checkpoints with the help of a validation set for each training run. We then merge the four single-task models using a scaling value $\lambda = 1$.

\section{Results and Discussion}
\subsection{Main Results: Scaling Multi-Task LLM Post-Training with \textsc{Omni-Think}} 

We evaluate \textsc{Omni-Thinker} across four diverse domains: Coding, Math, General QA, and Creative Writing, to assess how reinforcement learning with rule-based verifiable rewards and generative supervision supports multi-task generalization.
BWT matrix is computed following equation \ref{equ:BWT}, then Algorithm \ref{alg:constantBWT} is used to predict the accuracy of the model after curriculum learning, Appendix \ref{appendix:bwt_cur} for details.  The predicted best curriculum using Algorithm \ref{alg:lom} is Code $\rightarrow$ Math $\rightarrow$ QA $\rightarrow$ Writing while the worst is  Writing $\rightarrow$ QA $\rightarrow$ Math $\rightarrow$ Coding. 

Figure~\ref{fig:multitask_subplots} shows that Curriculum Multi-Task Learning with GRPO consistently yields the best results. Table~\ref{tab:appendix:main_results} further details how these gains vary by benchmarks.

In \textbf{Math}, Curriculum Learning (CL) achieves the highest average performance at 59.6\%, with the clearest gains on more complex reasoning tasks such as MinervaMath and OlympiadBench. These benchmarks benefit from strong rule-based reward signals and backward-transfer-guided task ordering. In contrast, datasets like AMC23 show minimal change because their relatively high baseline scores likely reflect smaller question sets and potential pretraining overlap rather than robust multi-step problem-solving.

In \textbf{General QA}, CL again performs best (52.2\%), followed by Model Merging (49.8\%) and Mixed Training GRPO (48.8\%). These improvements are driven by our Short-Form Open-Ended Supervision strategy: instead of generating responses in a fully open-ended and unconstrained fashion, the model is trained to produce complete answer strings given a diverse set of candidate responses, enabling the effective application of verifiable reward through simple string matching when training general-domain tasks.

For \textbf{Code Generation}, CL achieves 35.4\%, slightly ahead of Model Merging. Notably, we only evaluate on the subset of LiveCodeBench(24Oct-25Jan) problems released after Qwen2.5’s data cutoff, which ensures that these are unseen test items. This setup highlights CL’s significant generalization gains on novel problems, explaining the larger improvements on LiveCodeBench relative to static benchmarks like BigCodeBench, where data overlap is more likely. 

In \textbf{Creative Writing}, the introduction of our Long-Form Open-Ended Supervision strategy, employing the \textit{LLM-as-a-Judge} framework for pairwise evaluation, results in significant performance boosts (Curriculum-Guided at 84.2\% and Joint MT at 83.00\%), underscoring the advantage of our generative reward approach in subjective, open-ended tasks.

These results support our central hypothesis: The \textsc{Omni-Thinker} Framework, BWT-guided Curriculum Learning with hybrid rewards, enables a single unified policy to scale across structured and open-ended tasks alike, without relying on interleaving \textit{RLVR} on reasoning tasks and fine-tuning non-reasoning tasks.

\begin{table*}[ht]
    \centering
    \footnotesize
    \caption{Performance across benchmarks. \textbf{ST} = Single-Task RL (e.g., \textbf{ST Math} = RL trained only on math). 
    \textbf{MM} = Model Merging. 
    \textbf{JT} = Joint Training. 
    \textbf{CL} = Curriculum Learning. 
    Bolded values mark the best per row; underscored values mark the second best. Domains include Math (7 sets), MMLU-Pro (9 categories), Coding (2 sets), and Creative Writing (MT-Bench).}
    \resizebox{0.95\textwidth}{!}{%

    \begin{tabular}{ll|cccc|ccc|c}
    \toprule
    \textbf{Eval Task} 
    & \textbf{\shortstack{Base\\Model}}
    & \textbf{\shortstack{ST\\Coding}} 
    & \textbf{\shortstack{ST\\Math}} 
    & \textbf{\shortstack{ST\\QA}}
    & \textbf{\shortstack{ST\\Writing}}
    & \textbf{SFT} 
    & \textbf{MM} 
    & \textbf{JT} 
    & \textbf{CL$_{\text{best}}$} \\
    \midrule
    \multicolumn{10}{c}{\textbf{Math}} \\
    AIME24 & \textbf{18.0} & 13.3 & 14.7 & 14.0 & 15.3 & \underline{16.7} & 10.0 & 11.3 & 15.3 \\
    AMC23 & 57.5 & 57.5 & 60.0 & \underline{62.5} & 61.0 & 62.0 & 56.0 & 51.0 & \textbf{70.0} \\
    Gaokao2023en & 73.0 & 74.3 & 76.1 & 74.0 & 75.6 & 74.3 & 74.8 & \underline{76.6} & \textbf{77.1} \\
    MATH500 & 78.2 & 78.8 & \underline{80.4} & 75.4 & 79.2 & 76.8 & 79.8 & 77.6 & \textbf{81.0} \\
    MinervaMath & 64.3 & 64.0 & 66.5 & 63.2 & 61.8 & 65.1 & 66.2 & \underline{68.4} & \textbf{71.7} \\
    OlympiadBench & 42.1 & 43.0 & \underline{43.7} & 41.3 & 43.0 & 43.0 & 41.8 & 43.6 & \textbf{47.4} \\
    \textbf{Average} & 55.5 & 55.1 & \underline{56.9} & 55.1 & 56.0 & 56.3 & 54.8 & 54.8 & \textbf{60.4} \\
    \midrule
    \multicolumn{10}{c}{\textbf{General QA}} \\
    Biology & 57.6 & 56.8 & 52.3 & \underline{67.4} & 59.0 & 66.3 & 65.6 & 67.2 & \textbf{68.8} \\
    Business & 33.5 & 39.0 & 25.6 & \underline{58.7} & 33.0 & 48.2 & \textbf{59.8} & 49.8 & 47.5 \\
    Chemistry & 35.8 & 31.8 & 27.3 & \underline{47.7} & 38.3 & 44.1 & 42.5 & 42.1 & \textbf{50.7} \\
    CS & 53.7 & 48.1 & 50.2 & 55.1 & 52.0 & 53.7 & 53.9 & \underline{58.8} & \textbf{59.3} \\
    Economics & 42.7 & 49.2 & 38.7 & \textbf{62.9} & 44.9 & 59.6 & 62.0 & 56.8 & \underline{62.1} \\
    Engineering & 28.3 & 31.3 & 20.4 & \underline{37.5} & 26.6 & 37.8 & \textbf{38.1} & 35.8 & 37.1 \\
    Health & 46.7 & 46.2 & 45.2 & 51.0 & 47.2 & 45.7 & \underline{52.7} & 50.7 & \textbf{57.1} \\
    History & 37.3 & 33.3 & 34.7 & \underline{47.2} & 38.6 & 33.9 & \textbf{47.3} & 43.3 & 45.7 \\
    Law & 23.2 & 24.0 & 20.6 & \underline{27.9} & 23.3 & 26.8 & 26.6 & 27.5 & \textbf{29.7} \\
    Math & 55.4 & 52.6 & 50.4 & \underline{59.3} & 56.3 & 57.4 & 58.3 & 59.2 & \textbf{61.2} \\
    Other & 44.3 & 40.0 & 39.7 & 51.0 & 43.9 & 46.4 & \underline{51.8} & 49.9 & \textbf{53.3} \\
    Philosophy & 36.9 & 34.3 & 33.3 & \textbf{43.9} & 35.5 & 38.2 & 41.5 & 42.1 & \underline{42.9} \\
    Physics & 41.1 & 37.4 & 30.8 & \underline{53.7} & 41.6 & 49.8 & 46.7 & 48.0 & \textbf{55.6} \\
    Psychology & 50.9 & 51.5 & 45.4 & \underline{60.2} & 51.8 & 59.0 & 59.4 & 59.3 & \textbf{61.8} \\
    \textbf{Average} & 41.5 & 40.1 & 37.9 & \underline{51.3} & 42.0 & 47.8 & 49.8 & 48.8 & \textbf{52.2} \\
    \midrule
    \multicolumn{10}{c}{\textbf{Coding}} \\
    BigCodeBench & 46.5 & \textbf{50.4} & 46.7 & 47.1 & 46.8 & 44.5 & 48.1 & 47.2 & \underline{49.5} \\
    LiveCodeBench & 12.7 & \textbf{21.8} & 13.1 & 13.8 & 13.3 & 14.2 & 20.8 & 17.0 & \underline{21.3} \\
    \textbf{Average} & 29.6 & \textbf{36.1} & 29.9 & 30.5 & 30.1 & 29.3 & 34.4 & 32.1 & \underline{35.4} \\
    \midrule
    \multicolumn{10}{c}{\textbf{Creative Writing}} \\
    MT-Bench (Writing) & 74.2 & 71.6 & 74.2 & 63.0 & 78.3 & 44.2 & 67.5 & \underline{83.0} & \textbf{84.2} \\
    \bottomrule
    \end{tabular}
    }
    \vspace{-0.4cm}
    \label{tab:appendix:main_results}
\end{table*}

\subsection{Entropy Dynamics: Discussion}
\label{sec:entropy_discussion}

 Comparison between accuracies predictions to the actual test evaluation results for various curricula is depicted on Table \ref{table:test_pred_v_Test}. 
Predicted accuracies using test set backward transfers are surprisingly precise considering our simplifying assumptions, especially for the top curriculum.  We now discuss an identified cause of discrepancies.

We define the token-wise entropy of a policy $\pi_\theta$ on task $T_k$ 
ti measures the average per-token uncertainty of the policy across task samples.
\begin{equation}
    \mathcal E(\theta,T_k) := -\mathbb E_{q \sim \mathcal D_k, o\sim \pi_\theta^\otimes(\cdot\mid q)} \frac{1}{|o|}\sum_{t=1}^{|o|}\sum_{v\in \mathcal V}\pi_\theta(v\mid q,o_{<t})\log \pi_\theta(v\mid q,o_{<t}).
\end{equation}

\begin{wraptable}{r}{0.55\linewidth}
\vspace{-0.4cm}
\centering
\small
\caption{Test performance (\%) of single-task RL fine-tuning on General QA and Creative Writing respectively under different generation temperature (T) in training.}
\label{tab:ablation_temperature}
\begin{tabular}{l|cc|cc}
\toprule
\textbf{Eval Task} & 
\multicolumn{2}{c|}{\textbf{General QA}} & 
\multicolumn{2}{c}{\textbf{Creative Writing}} \\
 & T=1.0 & T=1.2 & T=1.0 & T=0.1 \\
\midrule
Math        & 55.1 & 54.8 & 55.6 & 57.8 \\
Coding      & 30.4 & 26.6 & 30.1 & 27.4 \\
General QA  & 51.4 & 13.9 & 42.0 & 44.9 \\
Creative Writing & 63.0 & 66.7 & 78.3 & 82.5 \\
\bottomrule
\end{tabular}
\vspace{-0.2cm}
\end{wraptable}

Entropy has been shown to drop during GRPO fine-tuning on reasoning, coding and more generally verified rewards \citep{rastogi2025magistral,cui2025entropy,yu2025dapo} with a correlation to higher accuracy until reaching a breaking point. 
Long training requires extra care towards entropy scaling either via regularization \citep{shen2025entropy} or dynamic temperature scaling.
Our multi-task setting differs in two key aspects compared to the above references.

First, we are using hybrid rewards with both verified and generative components.  The Creative Writing task is generative and is expected to increase entropy. Indeed, \citet{wang2025beyond} account for so-called {\it forking} tokens corresponding to structural choices of the output capturing most of the entropy. Reasoning tasks tend to require highly causal token sequences hence few forking tokens (low entropy) while generative tasks may allow more logical cuts at inference (higher entropy).  

Second, we train on multiple domains compared to mostly single-domain analysis in the references above. It is unclear a priori whether entropy decrease propagates from task to task.
\cite{agarwal2025unreasonable} show that fine-tuning to reduce entropy suffices to improve performance on multiple domains.
We hypothesize that models implicitly learn to emulate lower or higher temperatures as a mechanism to regulate entropy. 
In practice, the policy often produces logically flawed but rarely syntactically meaningless outputs, suggesting that its support lies within a constrained domain $V(q,o_{<t}) \subset \mathcal V$:  $
\sum_{v \in V(q,o_{<t})} \pi_\theta(v \mid q,o_{<t}) = 1$.
Scaling the final layer weights by a factor $\lambda < 1$ preserves this domain while increasing entropy, effectively raising the model’s base temperature. 
Such changes propagate across all tasks, not only the one being fine-tuned. 
Thus, even when two domains are sufficiently distinct for knowledge transfer to fail, the \textit{entropy dynamics} may still be measurable across tasks. 

\begin{table}[h]
\centering
\footnotesize
\setlength{\tabcolsep}{5pt}
\caption{Comparison of empirical and predicted test accuracies (\%). 
Each task column reports \textbf{Test} vs. \textbf{Predicted} accuracy for a given curriculum order. Standard deviations are rounded up.}
\label{table:test_pred_v_Test}
\begin{tabular}{c|cc|cc|cc|cc}
\toprule
\textbf{Curriculum} & 
\multicolumn{2}{c|}{\textbf{Math}} & 
\multicolumn{2}{c|}{\textbf{Coding}} & 
\multicolumn{2}{c|}{\textbf{QA}} & 
\multicolumn{2}{c}{\textbf{Writing}} \\
 & Test & Pred & Test & Pred & Test & Pred & Test & Pred \\
\midrule
$\mathcal{C}\,\mathcal{M}\,\mathcal{Q}\,\mathcal{W}$ & $60.4\!\pm\!0.3$ & $57.3$ & $35.4\!\pm\!0.3$ & $35.4$ & $52.2\!\pm\!0.1$ & $51.9$ & $84\!\pm\!2$ & $78.4$ \\
$\mathcal{Q}\,\mathcal{M}\,\mathcal{W}\,\mathcal{C}$ & $59.3\!\pm\!0.3$ & $55.6$ & $31.6\!\pm\!0.3$ & $36.7$ & $39.0\!\pm\!0.1$ & $38.8$ & $79\!\pm\!2$ & $78.4$ \\
$\mathcal{Q}\,\mathcal{W}\,\mathcal{C}\,\mathcal{M}$ & $60.4\!\pm\!0.3$ & $57.6$ & $31.9\!\pm\!0.3$ & $33.9$ & $36.3\!\pm\!0.1$ & $38.8$ & $82\!\pm\!2$ & $73.1$ \\
$\mathcal{W}\,\mathcal{Q}\,\mathcal{M}\,\mathcal{C}$ & $56.6\!\pm\!0.3$ & $55.5$ & $32.7\!\pm\!0.3$ & $36.1$ & $22.6\!\pm\!0.1$ & $38.4$ & $75\!\pm\!2$ & $62.1$ \\
\bottomrule
\end{tabular}
\end{table}

\subsection{Entropy Dynamics: Empirical Support}
\label{sec:entropy_support}
\begin{wrapfigure}{r}{.5\linewidth}
    \centering
    \vspace{-.5cm}
    \includegraphics[width=\linewidth]{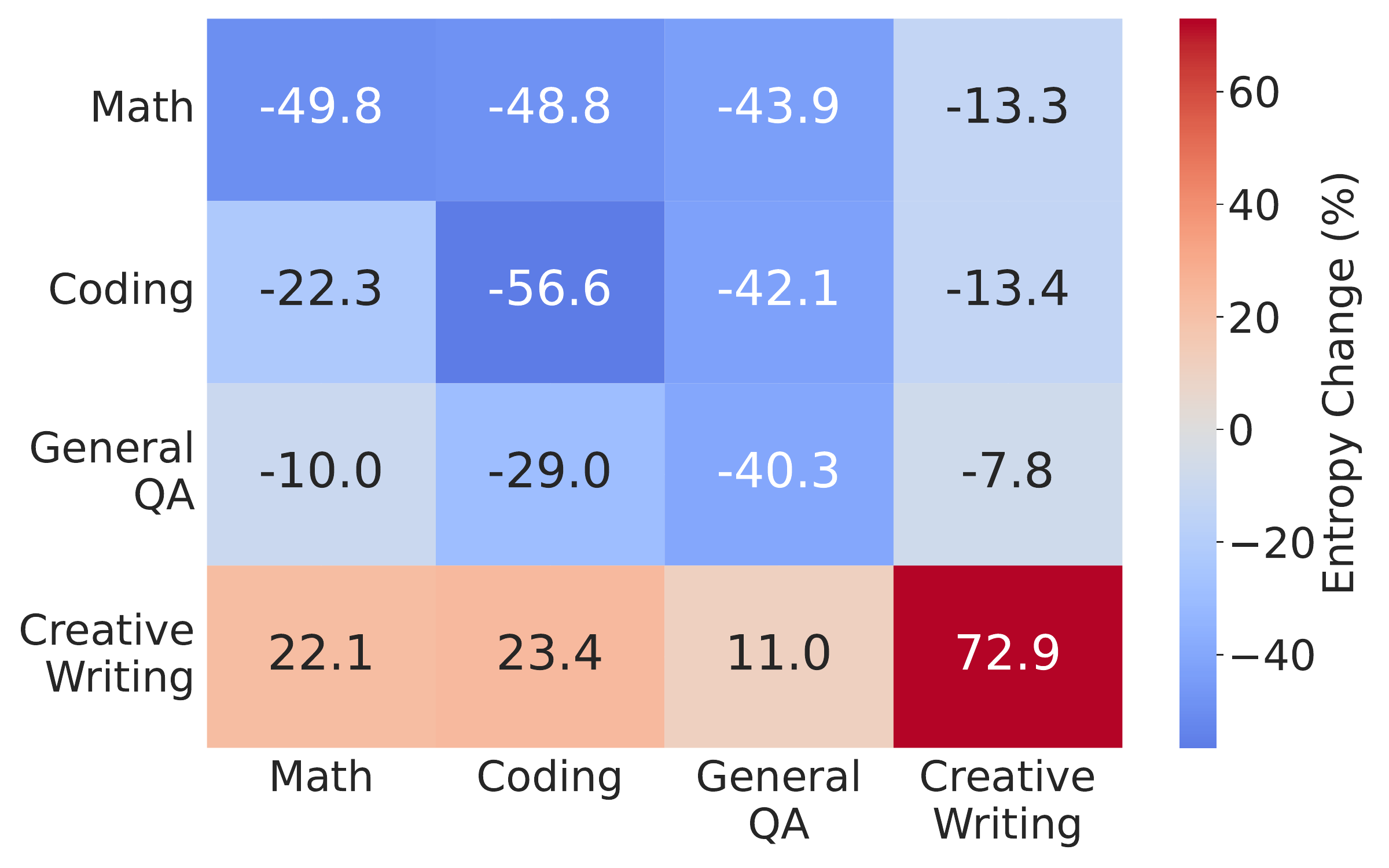}
    \caption{Validation Set Entropy Change Matrix.}
    \label{fig:entropy_bwt_heatmap}
    \vspace{-0.8cm}
\end{wrapfigure}
The intuitions laid out in the previous section are empirically supported by two experiments. 

We define the entropy change matrix as 
\begin{equation}
    H_{ij}:= \frac{\mathcal E(\theta(\theta_0,T_i),T_j)}{\mathcal E(\theta_0,T_j)}-1
\end{equation}
and compute it, see Figure \ref{fig:entropy_bwt_heatmap}. We observe that Math and Coding decrease entropy for all tasks (as previously observed for Verified Rewards) while Creative Writing increases entropy. Also, entropy change seems to depend primarily on the source task type, secondarily on the target task type, but not on their domain overlap.

We fine-tune the base model on QA and Writing task with different choices of temperature to emulate the effect of entropy modifications due to fine-tuning on entropy-increasing or entropy-decreasing tasks. 
On the one hand,  Writing is thought to benefits from temperature lowering coming from other tasks, we thus train the base model on writing with a lower temperature and expect the model to close the gap compared to the best Curriculum-trained model when evaluated with zero temperature.
On the other hand, QA  is thought to performs worse than expected in the worst curriculum due to the increased entropy coming from the trainig on Writing. We fine-tune the base model on QA with a higher temperature, then evaluate with zero temperature,  and expect QA performance to drop toward the low performance of the worse curriculum learning.  The results on table \ref{tab:ablation_temperature} shows that indeed the case.

\section{Related Work \& limitations}

\paragraph{Large Language Models and Multi-Task Learning} Early work like \citep{sanh2021multitask} showed that multi-task prompted training can encourage zero-shot generalization. \citet{dong2023abilities} further analyzed how mixing SFT data across domains can cause performance conflicts and forgetting, proposing Dual-stage Mixed Fine-tuning to alleviate these effects. However purely supervised objectives often encourage memorization rather than transferable reasoning. The Qwen3 model series \citep{yang2025qwen3} employs a four-stage post-training pipeline  in the order of reasoning, non-reasoning, and general-domain under a mix of supervised fine-tuning and reinforcement learning. In comparison, the post-training process for Command-A \citep{cohere2025command} alternates between training multiple expert models separately and merging the experts' parameters into a ``Soup Model'' during its SFT and RL steps, before the model undergoes a polishing phase of preference alignment.
In contrast, our work integrates multi-task learning directly into a single RL framework. Its backward-transfer-guided curriculum orders tasks from least to most forgettable, drawing on continual learning insights \citep{lopez2017gradient} to reduce interference and maintain stable cross-task performance.

\paragraph{Large Language Models and Reinforcement Learning}
Reinforcement Learning with Verified Rewards has demonstrated effectiveness for tasks with deterministic correctness signals such as math or code generation \citep{lambert2024t, shao2024deepseekmath, kimiteam2025kimik15scalingreinforcement, deepseekai2025deepseekr1incentivizing}. Recent frameworks like General-Reasoner~\citep{ma2025general}, Nemotron-Crossthink~\citep{crossthink} and X-REASONER~\citep{liu2025xreasonergeneralizablereasoning} expand this to broader reasoning by blending multi-domain corpora and structured answer templates. However, these tasks still largely remain largely confined to verifiable STEM problems or multiple-choice formats, leaving open-ended generation, such as creative writing, insufficiently addressed. To bridge this gap, \citet{su2025crossingrewardbridge} propose a generative
reward model (GRM) to replace rule-based signals. Although this improves RL and makes it applicable to general-domain QA when references exist, the approach is still restricted to  verifiable tasks.
In contrast, our approach integrates hybrid verifiable and preference-based rewards within a single RL loop, enabling consistent optimization across both structured and open-ended tasks. Moreover, our curriculum design, guided by backward transfer, helps maintain stable cross-task performance even for tasks lacking deterministic evaluation criteria.

\section{Conclusion}
We presented \textsc{Omni-Thinker}, a unified reinforcement learning framework that enables large language models to handle both structured and open-ended tasks under a single policy.  By combining rule-based verifiable rewards and generative preference-based supervision, our method improves generalization while mitigating forgetting and interference.
Our findings show that effective multi-task LLM post-training depends not only on reward design but also on how tasks are sequenced and optimized together. Ordering tasks from structured to open-ended domains based on backward transfer reduces forgetting and enhances cross-domain performance.
Overall, \textsc{Omni-Thinker} advances the goal of general-purpose LLMs that can learn from both verifiable and subjective feedback, bridging structured reasoning, open-ended question answering, and creative generation in a single post-training framework.
 
\paragraph{Limitations:}
Further work is needed to test this approach across a broader range of tasks and domains, including those that require logical reasoning over graph-structured data ~\citep{zhou2024enhancing} and knowledge-base retrieval ~\citep{dehghan2024ewek}, as well as more diverse open-ended domains such as mixed-initiative collaborative storytelling and co-creativity ~\citep{kreminski2024intent}.  
Our discussion on entropy is consistent with the existing literature however we lack 1) experiments backing our intuition that forking tokens are clearly identifiable and more frequent in generative tasks is lacking 2) a direct measurement of temperature scaling due to fine-tuning. 
Furthermore, our discussion on entropy is mostly qualitative. A quantitative study of entropy effects could allow to avoid failure cases of our accuracy predictions hence of our task scheduler. 
Finally, we fully leverage BWT only for curriculum learning. A comprehensive framework combining BWT, entropy and gradient norm in a quantitative manner in the continuous limit would allow for dynamical joint training with mixed batches.

\clearpage
\newpage
\bibliography{bibliography}

\begin{thebibliography}{48}
\providecommand{\natexlab}[1]{#1}
\providecommand{\url}[1]{\texttt{#1}}
\expandafter\ifx\csname urlstyle\endcsname\relax
  \providecommand{\doi}[1]{doi: #1}\else
  \providecommand{\doi}{doi: \begingroup \urlstyle{rm}\Url}\fi

\bibitem[Agarwal et~al.(2025)Agarwal, Zhang, Yuan, Han, and Peng]{agarwal2025unreasonable}
Shivam Agarwal, Zimin Zhang, Lifan Yuan, Jiawei Han, and Hao Peng.
\newblock {The unreasonable effectiveness of entropy minimization in llm reasoning}.
\newblock \emph{arXiv preprint arXiv:2505.15134}, 2025.

\bibitem[Akter et~al.(2025)Akter, Prabhumoye, Novikov, Han, Lin, Bakhturina, Nyberg, Choi, Patwary, Shoeybi, and Catanzaro]{crossthink}
Syeda~Nahida Akter, Shrimai Prabhumoye, Matvei Novikov, Seungju Han, Ying Lin, Evelina Bakhturina, Eric Nyberg, Yejin Choi, Mostofa Patwary, Mohammad Shoeybi, and Bryan Catanzaro.
\newblock {{N}emotron-{C}ross{T}hink: Scaling Self-Learning beyond Math Reasoning}, 2025.

\bibitem[Chen et~al.(2025)Chen, Hu, Zou, Wu, Wang, Hooi, and He]{chen2025judgelrm}
Nuo Chen, Zhiyuan Hu, Qingyun Zou, Jiaying Wu, Qian Wang, Bryan Hooi, and Bingsheng He.
\newblock {Judgelrm: Large reasoning models as a judge}.
\newblock \emph{arXiv preprint arXiv:2504.00050}, 2025.

\bibitem[Cohere et~al.(2025)Cohere, Ahmadian, Ahmed, Alammar, Alizadeh, Alnumay, Althammer, Arkhangorodsky, Aryabumi, Aumiller, et~al.]{cohere2025command}
Team Cohere, Arash Ahmadian, Marwan Ahmed, Jay Alammar, Milad Alizadeh, Yazeed Alnumay, Sophia Althammer, Arkady Arkhangorodsky, Viraat Aryabumi, Dennis Aumiller, et~al.
\newblock {Command a: An enterprise-ready large language model}.
\newblock \emph{arXiv preprint arXiv:2504.00698}, 2025.

\bibitem[Cui et~al.(2025)Cui, Zhang, Chen, Yuan, Wang, Zuo, Li, Fan, Chen, Chen, et~al.]{cui2025entropy}
Ganqu Cui, Yuchen Zhang, Jiacheng Chen, Lifan Yuan, Zhi Wang, Yuxin Zuo, Haozhan Li, Yuchen Fan, Huayu Chen, Weize Chen, et~al.
\newblock {The entropy mechanism of reinforcement learning for reasoning language models}.
\newblock \emph{arXiv preprint arXiv:2505.22617}, 2025.

\bibitem[Dehghan et~al.(2024)Dehghan, Alomrani, Bagga, Alfonso-Hermelo, Bibi, Ghaddar, Zhang, Li, Hao, Liu, et~al.]{dehghan2024ewek}
Mohammad Dehghan, Mohammad Alomrani, Sunyam Bagga, David Alfonso-Hermelo, Khalil Bibi, Abbas Ghaddar, Yingxue Zhang, Xiaoguang Li, Jianye Hao, Qun Liu, et~al.
\newblock {EWEK-QA: Enhanced Web and Efficient Knowledge Graph Retrieval for Citation-based Question Answering Systems}.
\newblock In \emph{Proceedings of the 62nd Annual Meeting of the Association for Computational Linguistics}, 2024.

\bibitem[Dong et~al.(2023)Dong, Yuan, Lu, Li, Xue, Liu, Wang, Yuan, Zhou, and Zhou]{dong2023abilities}
Guanting Dong, Hongyi Yuan, Keming Lu, Chengpeng Li, Mingfeng Xue, Dayiheng Liu, Wei Wang, Zheng Yuan, Chang Zhou, and Jingren Zhou.
\newblock {How abilities in large language models are affected by supervised fine-tuning data composition}.
\newblock \emph{arXiv preprint arXiv:2310.05492}, 2023.

\bibitem[Du et~al.(2025)Du, Yao, Ma, Wang, Zheng, Zhu, Liu, Liang, Jin, Wei, et~al.]{pteam2025supergpqascalingllmevaluation}
Xinrun Du, Yifan Yao, Kaijing Ma, Bingli Wang, Tianyu Zheng, King Zhu, Minghao Liu, Yiming Liang, Xiaolong Jin, Zhenlin Wei, et~al.
\newblock Supergpqa: Scaling llm evaluation across 285 graduate disciplines.
\newblock \emph{arXiv preprint arXiv:2502.14739}, 2025.

\bibitem[Dubey et~al.(2024)Dubey, Jauhri, Pandey, Kadian, Al-Dahle, Letman, Mathur, Schelten, Yang, Fan, et~al.]{dubey2024llama}
Abhimanyu Dubey, Abhinav Jauhri, Abhinav Pandey, Abhishek Kadian, Ahmad Al-Dahle, Aiesha Letman, Akhil Mathur, Alan Schelten, Amy Yang, Angela Fan, et~al.
\newblock {The llama 3 herd of models}.
\newblock \emph{arXiv preprint arXiv:2407.21783}, 2024.

\bibitem[Floudas \& Pardalos(2008)Floudas and Pardalos]{floudas2008encyclopedia}
Christodoulos~A Floudas and Panos~M Pardalos.
\newblock \emph{{Encyclopedia of optimization}}.
\newblock Springer Science \& Business Media, 2008.

\bibitem[Guo et~al.(2025)Guo, Yang, Zhang, Song, Wang, Zhu, Xu, Zhang, Ma, Bi, et~al.]{deepseekai2025deepseekr1incentivizing}
Daya Guo, Dejian Yang, Haowei Zhang, Junxiao Song, Peiyi Wang, Qihao Zhu, Runxin Xu, Ruoyu Zhang, Shirong Ma, Xiao Bi, et~al.
\newblock Deepseek-r1 incentivizes reasoning in llms through reinforcement learning.
\newblock \emph{Nature}, 2025.

\bibitem[He et~al.(2024)He, Luo, Bai, Hu, Thai, Shen, Hu, Han, Huang, Zhang, Liu, Qi, Liu, and Sun]{he2024olympiadbench}
Chaoqun He, Renjie Luo, Yuzhuo Bai, Shengding Hu, Zhen~Leng Thai, Junhao Shen, Jinyi Hu, Xu~Han, Yujie Huang, Yuxiang Zhang, Jie Liu, Lei Qi, Zhiyuan Liu, and Maosong Sun.
\newblock {{OlympiadBench}: A Challenging Benchmark for Promoting {AGI} with Olympiad-Level Bilingual Multimodal Scientific Problems}.
\newblock In \emph{Proceedings of the 62nd Annual Meeting of the Association for Computational Linguistics}, 2024.

\bibitem[Hendrycks et~al.(2021)Hendrycks, Burns, Kadavath, Arora, Basart, Tang, Song, and Steinhardt]{hendrycksmath2021}
Dan Hendrycks, Collin Burns, Saurav Kadavath, Akul Arora, Steven Basart, Eric Tang, Dawn Song, and Jacob Steinhardt.
\newblock {Measuring Mathematical Problem Solving With the {MATH} Dataset}.
\newblock In \emph{Thirty-fifth Conference on Neural Information Processing Systems Datasets and Benchmarks Track}, 2021.

\bibitem[HuggingFace(2025)]{openr1}
HuggingFace.
\newblock {Open R1: A fully open reproduction of DeepSeek-R1}, January 2025.
\newblock URL \url{https://github.com/huggingface/open-r1}.

\bibitem[Hurst et~al.(2024)Hurst, Lerer, Goucher, Perelman, Ramesh, Clark, Ostrow, Welihinda, Hayes, Radford, et~al.]{hurst2024gpt}
Aaron Hurst, Adam Lerer, Adam~P Goucher, Adam Perelman, Aditya Ramesh, Aidan Clark, AJ~Ostrow, Akila Welihinda, Alan Hayes, Alec Radford, et~al.
\newblock {Gpt-4o system card}.
\newblock \emph{arXiv preprint arXiv:2410.21276}, 2024.

\bibitem[Jain et~al.(2024)Jain, Han, Gu, Li, Yan, Zhang, Wang, Solar-Lezama, Sen, and Stoica]{jain2024livecodebenchholisticcontaminationfree}
Naman Jain, King Han, Alex Gu, Wen-Ding Li, Fanjia Yan, Tianjun Zhang, Sida Wang, Armando Solar-Lezama, Koushik Sen, and Ion Stoica.
\newblock {LiveCodeBench: Holistic and Contamination Free Evaluation of Large Language Models for Code}.
\newblock In \emph{The Thirteenth International Conference on Learning Representations}, 2024.

\bibitem[Kimi-Team et~al.(2025)Kimi-Team, Du, Gao, Xing, Jiang, Chen, Li, Xiao, Du, Liao, Tang, Wang, Zhang, Yuan, Lu, Tang, Sung, Wei, Lai, Guo, Zhu, Ding, Hu, Yang, Zhang, et~al.]{kimiteam2025kimik15scalingreinforcement}
Kimi-Team, Angang Du, Bofei Gao, Bowei Xing, Changjiu Jiang, Cheng Chen, Cheng Li, Chenjun Xiao, Chenzhuang Du, Chonghua Liao, Chuning Tang, Congcong Wang, Dehao Zhang, Enming Yuan, Enzhe Lu, Fengxiang Tang, Flood Sung, Guangda Wei, Guokun Lai, Haiqing Guo, Han Zhu, Hao Ding, Hao Hu, Hao Yang, Hao Zhang, et~al.
\newblock {Kimi k1.5: Scaling Reinforcement Learning with LLMs}, 2025.

\bibitem[Kreminski et~al.(2024)Kreminski, Chung, and Dickinson]{kreminski2024intent}
Max Kreminski, John Joon~Young Chung, and Melanie Dickinson.
\newblock {Intent Elicitation in Mixed-Initiative Co-Creativity.}
\newblock In \emph{IUI Workshops}, 2024.

\bibitem[Lambert et~al.(2024)Lambert, Morrison, Pyatkin, Huang, Ivison, Brahman, Miranda, Liu, Dziri, Lyu, et~al.]{lambert2024t}
Nathan Lambert, Jacob Morrison, Valentina Pyatkin, Shengyi Huang, Hamish Ivison, Faeze Brahman, Lester James~V Miranda, Alisa Liu, Nouha Dziri, Shane Lyu, et~al.
\newblock {Tulu 3: Pushing frontiers in open language model post-training}.
\newblock \emph{arXiv preprint arXiv:2411.15124}, 2024.

\bibitem[Lewkowycz et~al.(2022)Lewkowycz, Andreassen, Dohan, Dyer, Michalewski, Ramasesh, Slone, Anil, Schlag, Gutman-Solo, Wu, Neyshabur, Gur-Ari, and Misra]{lewkowycz2022solving}
Aitor Lewkowycz, Anders~Johan Andreassen, David Dohan, Ethan Dyer, Henryk Michalewski, Vinay~Venkatesh Ramasesh, Ambrose Slone, Cem Anil, Imanol Schlag, Theo Gutman-Solo, Yuhuai Wu, Behnam Neyshabur, Guy Gur-Ari, and Vedant Misra.
\newblock {Solving Quantitative Reasoning Problems with Language Models}.
\newblock In Alice~H. Oh, Alekh Agarwal, Danielle Belgrave, and Kyunghyun Cho (eds.), \emph{Advances in Neural Information Processing Systems}, 2022.

\bibitem[Liao et~al.(2024)Liao, Luo, Li, Wu, and Fan]{liao2024mario}
Minpeng Liao, Wei Luo, Chengxi Li, Jing Wu, and Kai Fan.
\newblock {MARIO: MAth Reasoning with code Interpreter Output--A Reproducible Pipeline}.
\newblock \emph{arXiv preprint arXiv:2401.08190}, 2024.

\bibitem[Liu et~al.(2024)Liu, Feng, Xue, Wang, Wu, Lu, Zhao, Deng, Zhang, Ruan, et~al.]{liu2024deepseek}
Aixin Liu, Bei Feng, Bing Xue, Bingxuan Wang, Bochao Wu, Chengda Lu, Chenggang Zhao, Chengqi Deng, Chenyu Zhang, Chong Ruan, et~al.
\newblock {Deepseek-v3 technical report}.
\newblock \emph{arXiv preprint arXiv:2412.19437}, 2024.

\bibitem[Liu \& Zhang(2025)Liu and Zhang]{code-r1}
Jiawei Liu and Lingming Zhang.
\newblock {Code-R1: Reproducing R1 for Code with Reliable Rewards}.
\newblock 2025.

\bibitem[Liu et~al.(2025)Liu, Zhang, Qin, Ossowski, Gu, Jin, Kiblawi, Preston, Wei, Vozila, Naumann, and Poon]{liu2025xreasonergeneralizablereasoning}
Qianchu Liu, Sheng Zhang, Guanghui Qin, Timothy Ossowski, Yu~Gu, Ying Jin, Sid Kiblawi, Sam Preston, Mu~Wei, Paul Vozila, Tristan Naumann, and Hoifung Poon.
\newblock {X-Reasoner: Towards Generalizable Reasoning Across Modalities and Domains}.
\newblock \emph{arXiv preprint arXiv:2505.03981}, 2025.

\bibitem[Lopez-Paz \& Ranzato(2017)Lopez-Paz and Ranzato]{lopez2017gradient}
David Lopez-Paz and Marc'Aurelio Ranzato.
\newblock {Gradient episodic memory for continual learning}.
\newblock \emph{Advances in neural information processing systems}, 30, 2017.

\bibitem[Luo et~al.(2025)Luo, Tan, Wong, Shi, Tang, Roongta, Cai, Luo, Li, Popa, and Stoica]{deepscaler2025SurpassingO1preview}
Michael Luo, Sijun Tan, Justin Wong, Xiaoxiang Shi, William~Y. Tang, Manan Roongta, Colin Cai, Jeffrey Luo, Li~Erran Li, Raluca~Ada Popa, and Ion Stoica.
\newblock {DeepScaleR: Surpassing O1-Preview with a 1.5B Model by Scaling RL}, 2025.

\bibitem[Ma et~al.(2025)Ma, Liu, Jiang, Zhang, Ma, and Chen]{ma2025general}
Xueguang Ma, Qian Liu, Dongfu Jiang, Ge~Zhang, Zejun Ma, and Wenhu Chen.
\newblock {General-reasoner: Advancing llm reasoning across all domains}.
\newblock \emph{arXiv preprint arXiv:2505.14652}, 2025.

\bibitem[{MAA}(2023)]{MAA2023_AMC}
{MAA}.
\newblock {American mathematics competitions}.
\newblock \url{https://maa.org/student-programs/amc/}, 2023.

\bibitem[{MAA}(2024)]{MAA2024_AIME}
{MAA}.
\newblock {American invitational mathematics examination}.
\newblock \url{https://maa.org/maa-invitational-competitions/}, 2024.

\bibitem[Nitral‑AI(2024)]{nitral2025creative}
Nitral‑AI.
\newblock {Creative\_Writing‑ShareGPT}.
\newblock \url{https://huggingface.co/datasets/Nitral-AI/Creative\_Writing-ShareGPT}, 2024.
\newblock Dataset.

\bibitem[Rastogi et~al.(2025)Rastogi, Jiang, Lo, Berrada, Lample, Rute, Barmentlo, Yadav, Khandelwal, Chandu, et~al.]{rastogi2025magistral}
Abhinav Rastogi, Albert~Q Jiang, Andy Lo, Gabrielle Berrada, Guillaume Lample, Jason Rute, Joep Barmentlo, Karmesh Yadav, Kartik Khandelwal, Khyathi~Raghavi Chandu, et~al.
\newblock {Magistral}.
\newblock \emph{arXiv preprint arXiv:2506.10910}, 2025.

\bibitem[Sanh et~al.(2021)Sanh, Webson, Raffel, Bach, Sutawika, Alyafeai, Chaffin, Stiegler, Scao, Raja, et~al.]{sanh2021multitask}
Victor Sanh, Albert Webson, Colin Raffel, Stephen~H Bach, Lintang Sutawika, Zaid Alyafeai, Antoine Chaffin, Arnaud Stiegler, Teven~Le Scao, Arun Raja, et~al.
\newblock {Multitask prompted training enables zero-shot task generalization}.
\newblock \emph{arXiv preprint arXiv:2110.08207}, 2021.

\bibitem[Schulman et~al.(2017)Schulman, Wolski, Dhariwal, Radford, and Klimov]{schulman2017proximal}
John Schulman, Filip Wolski, Prafulla Dhariwal, Alec Radford, and Oleg Klimov.
\newblock {Proximal policy optimization algorithms}.
\newblock \emph{arXiv preprint arXiv:1707.06347}, 2017.

\bibitem[Shao et~al.(2024)Shao, Wang, Zhu, Xu, Song, Bi, Zhang, Zhang, Li, Wu, et~al.]{shao2024deepseekmath}
Zhihong Shao, Peiyi Wang, Qihao Zhu, Runxin Xu, Junxiao Song, Xiao Bi, Haowei Zhang, Mingchuan Zhang, YK~Li, Y~Wu, et~al.
\newblock {Deepseekmath: Pushing the limits of mathematical reasoning in open language models}.
\newblock \emph{arXiv preprint arXiv:2402.03300}, 2024.

\bibitem[Shen(2025)]{shen2025entropy}
Han Shen.
\newblock {On Entropy Control in LLM-RL Algorithms}.
\newblock \emph{arXiv preprint arXiv:2509.03493}, 2025.

\bibitem[Su et~al.(2025)Su, Yu, Song, Li, Mi, Tu, Zhang, and Yu]{su2025crossingrewardbridge}
Yi~Su, Dian Yu, Linfeng Song, Juntao Li, Haitao Mi, Zhaopeng Tu, Min Zhang, and Dong Yu.
\newblock {Crossing the Reward Bridge: Expanding RL with Verifiable Rewards Across Diverse Domains}, 2025.

\bibitem[Wang et~al.(2025)Wang, Yu, Gao, Zheng, Liu, Lu, Dang, Chen, Yang, Zhang, et~al.]{wang2025beyond}
Shenzhi Wang, Le~Yu, Chang Gao, Chujie Zheng, Shixuan Liu, Rui Lu, Kai Dang, Xionghui Chen, Jianxin Yang, Zhenru Zhang, et~al.
\newblock {Beyond the 80/20 rule: High-entropy minority tokens drive effective reinforcement learning for llm reasoning}.
\newblock \emph{arXiv preprint arXiv:2506.01939}, 2025.

\bibitem[Wang et~al.(2024)Wang, Ma, Zhang, Ni, Chandra, Guo, Ren, Arulraj, He, Jiang, et~al.]{wang2024mmlupro}
Yubo Wang, Xueguang Ma, Ge~Zhang, Yuansheng Ni, Abhranil Chandra, Shiguang Guo, Weiming Ren, Aaran Arulraj, Xuan He, Ziyan Jiang, et~al.
\newblock Mmlu-pro: A more robust and challenging multi-task language understanding benchmark.
\newblock In \emph{Advances in Neural Information Processing Systems}, 2024.

\bibitem[Wu et~al.(2025)Wu, Yao, Liu, Liu, Fu, Han, Li, Zhen, Zhong, and Yuan]{yadav2023ties}
Han Wu, Yuxuan Yao, Shuqi Liu, Zehua Liu, Xiaojin Fu, Xiongwei Han, Xing Li, Hui-Ling Zhen, Tao Zhong, and Mingxuan Yuan.
\newblock {Unlocking Efficient Long-to-Short LLM Reasoning with Model Merging}, 2025.

\bibitem[Yadav et~al.(2023)Yadav, Tam, Choshen, Raffel, and Bansal]{yadav2023tiesmergingresolvinginterferencemerging}
Prateek Yadav, Derek Tam, Leshem Choshen, Colin~A Raffel, and Mohit Bansal.
\newblock Ties-merging: Resolving interference when merging models.
\newblock In \emph{Advances in Neural Information Processing Systems}, 2023.

\bibitem[Yang et~al.(2024)Yang, Yang, Zhang, Hui, Zheng, Yu, Li, Liu, Huang, Wei, et~al.]{yang2024qwen2}
An~Yang, Baosong Yang, Beichen Zhang, Binyuan Hui, Bo~Zheng, Bowen Yu, Chengyuan Li, Dayiheng Liu, Fei Huang, Haoran Wei, et~al.
\newblock {Qwen2. 5 Technical Report}.
\newblock \emph{arXiv e-prints}, 2024.

\bibitem[Yang et~al.(2025)Yang, Li, Yang, Zhang, Hui, Zheng, Yu, Gao, Huang, Lv, et~al.]{yang2025qwen3}
An~Yang, Anfeng Li, Baosong Yang, Beichen Zhang, Binyuan Hui, Bo~Zheng, Bowen Yu, Chang Gao, Chengen Huang, Chenxu Lv, et~al.
\newblock {Qwen3 technical report}.
\newblock \emph{arXiv preprint arXiv:2505.09388}, 2025.

\bibitem[Yu et~al.(2025)Yu, Zhang, Zhu, Yuan, Zuo, Yue, Dai, Fan, Liu, Liu, et~al.]{yu2025dapo}
Qiying Yu, Zheng Zhang, Ruofei Zhu, Yufeng Yuan, Xiaochen Zuo, Yu~Yue, Weinan Dai, Tiantian Fan, Gaohong Liu, Lingjun Liu, et~al.
\newblock {Dapo: An open-source llm reinforcement learning system at scale}, 2025.

\bibitem[Yuan et~al.(2023)Yuan, Yuan, Li, Dong, Lu, Tan, Zhou, and Zhou]{yuan2023scaling}
Zheng Yuan, Hongyi Yuan, Chengpeng Li, Guanting Dong, Keming Lu, Chuanqi Tan, Chang Zhou, and Jingren Zhou.
\newblock {Scaling relationship on learning mathematical reasoning with large language models}.
\newblock \emph{arXiv preprint arXiv:2308.01825}, 2023.

\bibitem[Zhang et~al.(2025)Zhang, Hosseini, Bansal, Kazemi, Kumar, and Agarwal]{zhang2024generativerm}
Lunjun Zhang, Arian Hosseini, Hritik Bansal, Mehran Kazemi, Aviral Kumar, and Rishabh Agarwal.
\newblock {Generative verifiers: Reward modeling as next-token prediction}.
\newblock In \emph{The Thirteenth International Conference on Learning Representations}, 2025.

\bibitem[Zheng et~al.(2023)Zheng, Chiang, Sheng, Zhuang, Wu, Zhuang, Lin, Li, Li, Xing, et~al.]{zheng2023judging}
Lianmin Zheng, Wei-Lin Chiang, Ying Sheng, Siyuan Zhuang, Zhanghao Wu, Yonghao Zhuang, Zi~Lin, Zhuohan Li, Dacheng Li, Eric Xing, et~al.
\newblock {Judging llm-as-a-judge with mt-bench and chatbot arena}.
\newblock \emph{Advances in Neural Information Processing Systems}, 36, 2023.

\bibitem[Zhou et~al.(2024)Zhou, Ghaddar, Zhang, Ma, Hu, Pal, Coates, Wang, Zhang, and Hao]{zhou2024enhancing}
Jiaming Zhou, Abbas Ghaddar, Ge~Zhang, Liheng Ma, Yaochen Hu, Soumyasundar Pal, Mark Coates, Bin Wang, Yingxue Zhang, and Jianye Hao.
\newblock {Enhancing logical reasoning in large language models through graph-based synthetic data}.
\newblock \emph{arXiv preprint arXiv:2409.12437}, 2024.

\bibitem[Zhuo et~al.(2024)Zhuo, Vu, Chim, Hu, Yu, Widyasari, Yusuf, Zhan, He, Paul, et~al.]{zhuo2024bigcodebench}
Terry~Yue Zhuo, Minh~Chien Vu, Jenny Chim, Han Hu, Wenhao Yu, Ratnadira Widyasari, Imam Nur~Bani Yusuf, Haolan Zhan, Junda He, Indraneil Paul, et~al.
\newblock {Bigcodebench: Benchmarking code generation with diverse function calls and complex instructions}.
\newblock \emph{arXiv preprint arXiv:2406.15877}, 2024.

\end{thebibliography}
\bibliographystyle{iclr2025_conference}

\clearpage
\newpage
\appendix
\appendix
\section{Appendix}
\subsection{Reward Estimation}
Omni-Thinker employs a hybrid reward system combining rule-based correctness (math, code, QA) with preference-based supervision (creative writing) in a unified RL framework. We define task-specific reward functions as $R_k(q, o)$, where $o$ denotes the model output and $q$ is the prompt provided to the model. Each reward function captures domain-relevant correctness criteria, assessing whether $o$ satisfies symbolic constraints, passes execution tests, or is preferred over alternatives under subjective evaluation. While some rewards (e.g., math and code) are strictly deterministic, others, such as LLM-as-a-Judge comparisons, are inherently stochastic but executed at low decoding temperature to ensure stable and consistent supervision. All reward functions are designed to be domain-aware and automatable, supporting scalable reinforcement learning across both structured and generative tasks.

\paragraph{Primary Rewards.}
Each task employs a tailored correctness criterion:
\begin{itemize}
    \item \textbf{Math:}
    Elements of the math dataset are couples $(q,a_q)$ where $q$ is a prompt and $a_q$ is a token sequence stating the answer. We implement a $\texttt{verify}_{\texttt{math}}(o,a)$  function that combines regular expression and symbolic parser to checks that $a$ (or an equivalent acceptable answer) is in $o$ within tags  \texttt{<answer>}. More formally
    \[
    r_{\texttt{math}}(q, o) = \mathbbm{1}\left\{\texttt{verify}_{\texttt{math}}(o, a_q) = \texttt{true} \right\}\,.
    \]
    
    \item \textbf{Code Generation:}
    Each element of the code dataset is a tuple $(q, \texttt{unittest}_q, \texttt{test\_case}_q)$ where $q$ is a prompt, $\texttt{unittest}_q$  is a unit test function and $\texttt{test\_case}_q$ is a set of test cases. 
     Given an output $o$ for prompt $q$, the generated code $o_{\texttt{ans}}$ is extracted from the output $o$ using regular expressions, the unit test $\texttt{unittest}_q(o_{\texttt{ans}},x)$ is executed in a sandboxed environment for every test case $x\in  \texttt{test\_case}_q$.
    More formally
    \[
    r_{\texttt{code}}(q,o) = \prod_{x \in \texttt{test\_case}_q}\mathbbm{1}\left\{ \texttt{unittest}_q(o_{\texttt{ans}},x) \right\}
    \]

    \item \textbf{General QA:}
    Each element of the dataset for General QA is a couple $(q,a_q)$ of prompt and answer. The reward is defined by extracting the answer $o_{\texttt{ans}}$ from the output $o$ using regular expressions and testing it against the ground truth $a$. More formally:
    \[
    r_{\texttt{qa}}(q,o) = \mathbbm{1}\left\{ o_{\texttt{ans}} = a_q \right\}
    \]
    which returns 1 if the predicted answer matches the ground-truth string exactly.
    
    \item \textbf{Creative Writing:}
    Each element of the dataset for Creative writing is a couple $(q,o_{\texttt{ref},q})$. Given an output $o$ on prompt $q$, the reward is computed by calling a fixed \textit{LLM-as-a-Judge} model prompted to do a pairwise comparison between $o$ and $o_{\texttt{ref},q}$. More formally:
    \[
    r_{\text{writing}}(q,o) =
    \begin{cases}
    1.0 & \text{if } o \succ_q o_{\texttt{ref},q} \\
    0.5 & \text{if } o \sim_q o_{\texttt{ref},q} \\
    0.0 & \text{if } o \prec_q o_{\texttt{ref},q}
    \end{cases}
    \]
    where $ A \succ_q B$ means that the fixed \textit{LLM-as-a-Judge} model prefered $A$ over $B$ for request $q$, and $ A \sim_q B$ means it judges the answers as tied.

\end{itemize}

\paragraph{Auxiliary Rewards.}
To encourage structured outputs, we define formatting-based rewards shared across tasks:
\[
r_{\texttt{format}}(q,o) = \mathbbm{1}\left\{ \texttt{tags\_valid}(o) \right\} 
\]

\[
r_{\texttt{tags}}(q,o) = \frac{1}{4} \cdot \left| \texttt{tags\_present}(o) \right|
\]

Here, \texttt{tags\_valid} ensures proper nesting of \texttt{<think>} and \texttt{<answer>} tags, while \texttt{tags\_present} counts required structural markers.

\paragraph{Total Reward.}
We define the total reward as a weighted sum over both primary and auxiliary reward components. Let \( \mathcal{F}_k = \{r^{(1)}_k, r^{(2)}_k, \dots, r^{(m)}_k\} \) denote the set of reward functions associated with task \( k \), where each \( r^{(j)}_k \) measures a different aspect of correctness. Given a model output \( o \) and its associated evaluation context \( \phi_k \), the total reward is computed as:
\[
R_k(q,o) = \sum_{r \in \mathcal{F}_k} w_r \cdot r(q,o),
\]
where \( w_r \in [0,1] \) denotes the task-specific weight for the reward component \( r \). If a component reward is undefined, e.g., due to malformed or unparsable output, it is omitted from the sum. Samples with no valid reward components are excluded from policy updates.

\subsection{Results}

\subsubsection{Output Format Matters: Full-Text Answers Enhance Generalization}
We examine how output format impacts generalization by comparing models trained to generate full-text answers versus selecting letter choices in multiple-choice QA (MCQ). Using GRPO, we train two single-task policies on the training set, one prompted to produce full-text final answers at the end of its chain-of-thought completions, and the other to output only letter choices (e.g., “A”, “B”, “C”). 

\begin{wrapfigure}{r}{0.4\textwidth} 
  \centering
  \includegraphics[width=\linewidth]{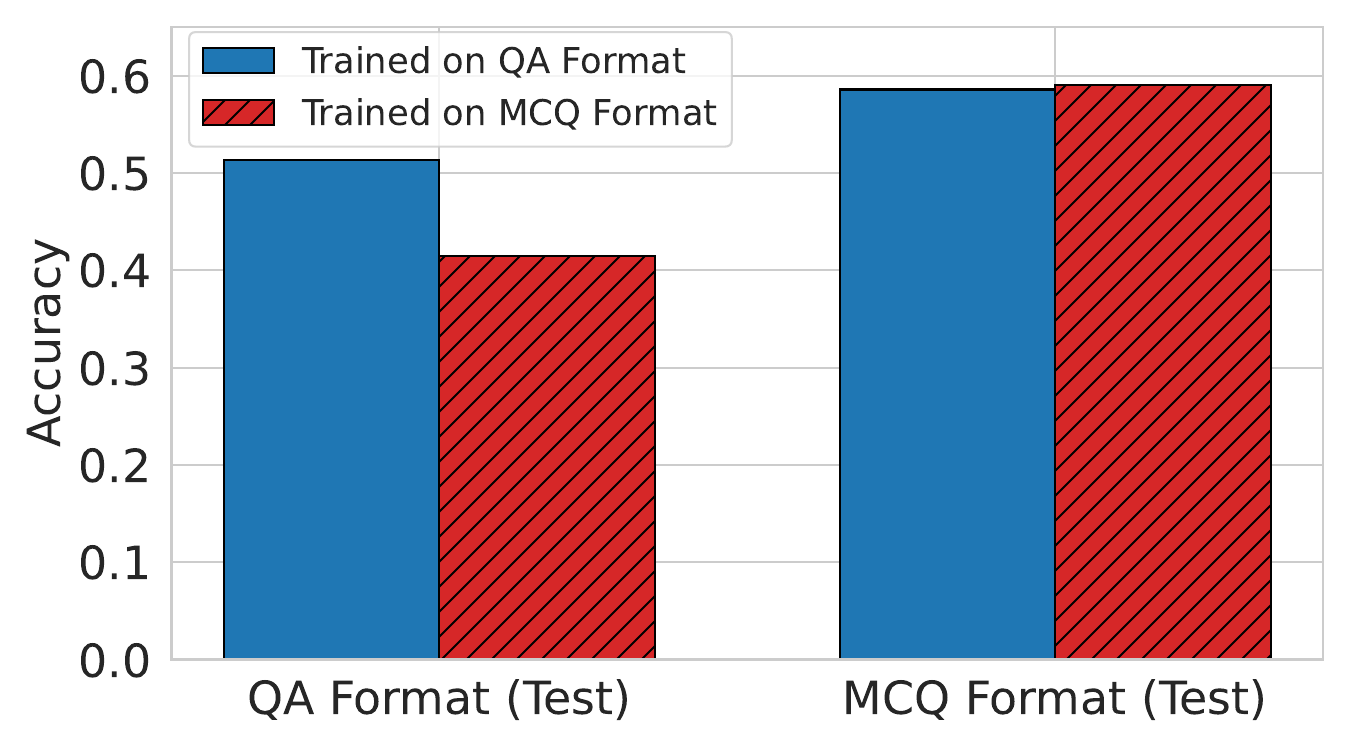}
  \caption{Models trained to generate full-text answers outperform those trained to select letter choices. 
  This format promotes deeper semantic understanding over shallow pattern matching or guessing.}
  \label{fig:mcq}
\end{wrapfigure}

As shown in Figure~\ref{fig:mcq}, the model trained to output full-text answers achieves significantly better generalization when evaluated with free-form QA prompts on MMLU Pro (51\% vs. 41\%). While the letter-choice model slightly outperforms when evaluated strictly on MCQ prompts, the full-text model remains competitive across both prompt formats.

These results suggest that training with complete, semantically grounded answers encourages deeper reasoning, improving the model’s ability to generalize beyond the specific format seen during training. In contrast, letter-choice training risks overfitting to shallow pattern matching, reducing transferability to realistic QA settings that often require articulated responses.

\subsubsection{Curriculum Accuracies Predictions}
\label{appendix:bwt_cur}
\begin{figure*}[h]
    \centering
    \begin{subfigure}[t]{0.45\textwidth}
        \centering
        \includegraphics[width=\linewidth]{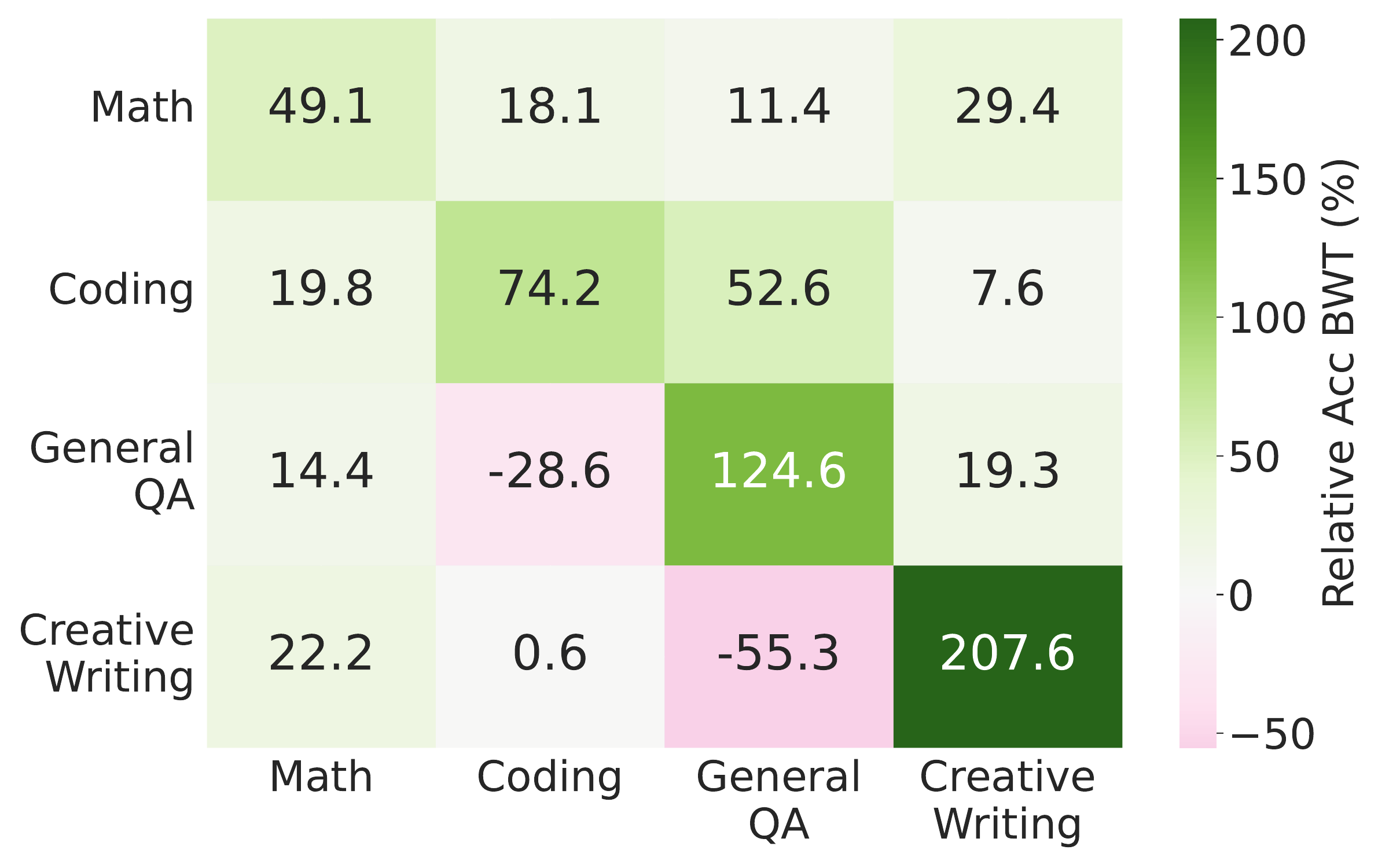}
        \label{fig:bwt_accruacy}
    \end{subfigure}
    \hfill
    \begin{subfigure}[t]{0.45\textwidth}
        \centering
        \includegraphics[width=\linewidth]{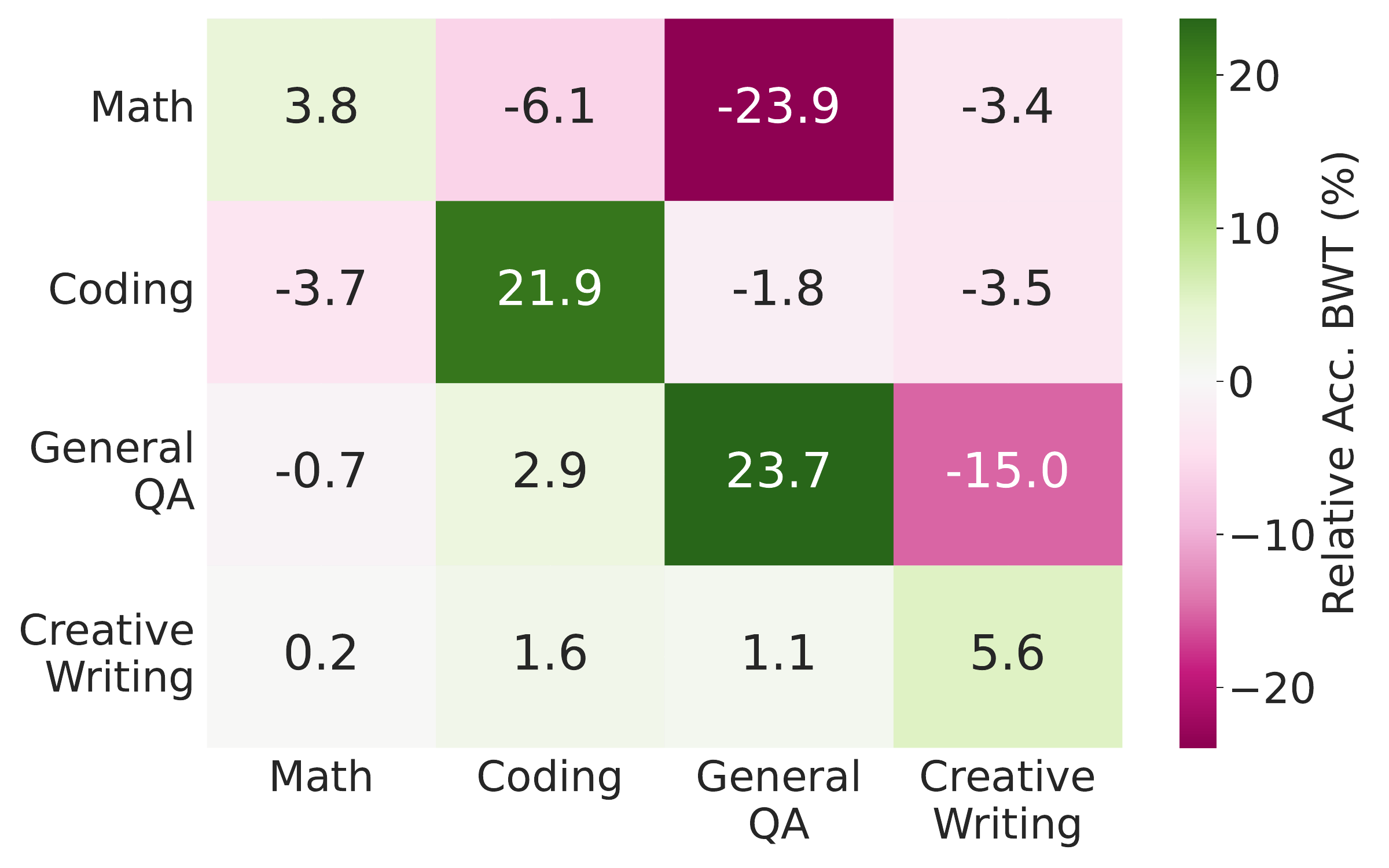}
        \label{fig:accuracy_bwt_heatmap}
    \end{subfigure}

    \caption{Comparison of backward transfer on validation set (left) and test set (right).}
    \label{fig:validation_vs_test_bwt}
\end{figure*}

On Figure \ref{fig:validation_vs_test_bwt} are provided BWT matrices for validation sets and test sets. We compute the ranking of curricula using both, see table \ref{tab:curriculum_ordered}. We observe that the Forgettability heuristic provides the second best prediction on test BWT, but exact LOM yields a potentially better candidate. Using validation BWT, the curriculum Code-Math-QA-Writing comes second,, but the proposed exact LOM should be rejected based on our entropy discussion.

\begin{table}[h]
\centering
\footnotesize
\caption{Curricula ranked by final geometrical average relative improvement 
$\overline{\Delta_\mathrm{rel}} := \frac{1}{K}\log{\prod_{k=1}^K\frac{\mathrm{Acc}(\theta_{s_\mathrm{max}},T_k)}{\mathrm{Acc}(\theta_{0},T_k)}}$
 on test set (left block) and eval set (right block), with predicted accuracies (\%) for each task at each order, and their ranking scores.}
\label{tab:curriculum_ordered}
\begin{tabular}{lccccc|lccccc}
\toprule
\multicolumn{6}{c|}{\textbf{Test set order }} & \multicolumn{6}{c}{\textbf{Validation set order}} \\
Order & M & C & Q & W & $\overline{\Delta_\mathrm{rel}}$ & Order & M & C & Q & W & $\overline{\Delta_\mathrm{rel}}$ \\
\midrule
$\mathcal{M}\,\mathcal{C}\,\mathcal{Q}\,\mathcal{W}$ & 55.2 & 37.7 & 51.9 & 78.4 & 12.85 & $\mathcal{C}\,\mathcal{W}\,\mathcal{M}\,\mathcal{Q}$ & 34.9 & 42.5 & 26.8 & 35.4 & 26.00 \\
$\mathcal{C}\,\mathcal{M}\,\mathcal{Q}\,\mathcal{W}$ & 57.3 & 35.4 & 51.9 & 78.4 & 12.22 & $\mathcal{C}\,\mathcal{M}\,\mathcal{Q}\,\mathcal{W}$ & 36.7 & 42.5 & 27.5 & 32.6 & 25.85 \\
$\mathcal{M}\,\mathcal{Q}\,\mathcal{C}\,\mathcal{W}$ & 55.2 & 36.7 & 51.0 & 78.4 & 11.70 & $\mathcal{C}\,\mathcal{Q}\,\mathcal{W}\,\mathcal{M}$ & 33.1 & 42.5 & 28.1 & 35.3 & 25.81 \\
$\mathcal{M}\,\mathcal{Q}\,\mathcal{W}\,\mathcal{C}$ & 55.2 & 36.1 & 51.0 & 75.7 & 10.43 & $\mathcal{W}\,\mathcal{C}\,\mathcal{M}\,\mathcal{Q}$ & 34.9 & 44.7 & 26.8 & 33.4 & 25.77 \\
$\mathcal{M}\,\mathcal{C}\,\mathcal{W}\,\mathcal{Q}$ & 55.2 & 37.7 & 51.3 & 66.6 & 8.49 & $\mathcal{C}\,\mathcal{M}\,\mathcal{W}\,\mathcal{Q}$ & 36.7 & 42.5 & 26.8 & 32.8 & 25.29 \\
$\mathcal{C}\,\mathcal{M}\,\mathcal{W}\,\mathcal{Q}$ & 57.3 & 35.4 & 51.3 & 66.6 & 7.86 & $\mathcal{C}\,\mathcal{W}\,\mathcal{Q}\,\mathcal{M}$ & 33.1 & 42.5 & 27.4 & 35.4 & 25.25 \\
$\mathcal{M}\,\mathcal{W}\,\mathcal{C}\,\mathcal{Q}$ & 55.2 & 37.1 & 51.3 & 64.3 & 7.22 & $\mathcal{C}\,\mathcal{Q}\,\mathcal{M}\,\mathcal{W}$ & 34.8 & 42.5 & 28.1 & 32.6 & 25.09 \\
$\mathcal{C}\,\mathcal{W}\,\mathcal{M}\,\mathcal{Q}$ & 57.2 & 35.4 & 51.3 & 64.3 & 6.96 & $\mathcal{W}\,\mathcal{C}\,\mathcal{Q}\,\mathcal{M}$ & 33.1 & 44.7 & 27.4 & 33.4 & 25.02 \\
$\mathcal{W}\,\mathcal{M}\,\mathcal{C}\,\mathcal{Q}$ & 55.1 & 37.1 & 51.3 & 62.1 & 6.32 & $\mathcal{M}\,\mathcal{C}\,\mathcal{Q}\,\mathcal{W}$ & 37.0 & 40.2 & 27.5 & 32.6 & 24.71 \\
$\mathcal{M}\,\mathcal{W}\,\mathcal{Q}\,\mathcal{C}$ & 55.2 & 36.1 & 50.4 & 64.3 & 6.07 & $\mathcal{W}\,\mathcal{M}\,\mathcal{C}\,\mathcal{Q}$ & 35.3 & 42.3 & 26.8 & 33.4 & 24.64 \\
$\mathcal{W}\,\mathcal{C}\,\mathcal{M}\,\mathcal{Q}$ & 57.2 & 34.8 & 51.3 & 62.1 & 5.69 & $\mathcal{M}\,\mathcal{C}\,\mathcal{W}\,\mathcal{Q}$ & 37.0 & 40.2 & 26.8 & 32.8 & 24.16 \\
$\mathcal{C}\,\mathcal{Q}\,\mathcal{M}\,\mathcal{W}$ & 57.7 & 35.4 & 39.5 & 78.4 & 5.56 & $\mathcal{M}\,\mathcal{W}\,\mathcal{C}\,\mathcal{Q}$ & 37.0 & 42.3 & 26.8 & 30.9 & 23.93 \\
$\mathcal{W}\,\mathcal{M}\,\mathcal{Q}\,\mathcal{C}$ & 55.1 & 36.1 & 50.4 & 62.1 & 5.18 & $\mathcal{Q}\,\mathcal{C}\,\mathcal{W}\,\mathcal{M}$ & 33.1 & 41.3 & 26.1 & 35.3 & 23.23 \\
$\mathcal{Q}\,\mathcal{M}\,\mathcal{C}\,\mathcal{W}$ & 55.6 & 36.7 & 38.8 & 78.4 & 5.05 & $\mathcal{Q}\,\mathcal{W}\,\mathcal{C}\,\mathcal{M}$ & 33.1 & 43.4 & 26.1 & 33.3 & 23.00 \\
$\mathcal{C}\,\mathcal{Q}\,\mathcal{W}\,\mathcal{M}$ & 57.6 & 35.4 & 39.5 & 75.7 & 4.67 & $\mathcal{Q}\,\mathcal{C}\,\mathcal{M}\,\mathcal{W}$ & 34.8 & 41.3 & 26.1 & 32.6 & 22.52 \\
$\mathcal{Q}\,\mathcal{C}\,\mathcal{M}\,\mathcal{W}$ & 57.7 & 34.4 & 38.8 & 78.4 & 4.41 & $\mathcal{W}\,\mathcal{Q}\,\mathcal{C}\,\mathcal{M}$ & 33.1 & 43.4 & 25.4 & 33.4 & 22.45 \\
$\mathcal{Q}\,\mathcal{M}\,\mathcal{W}\,\mathcal{C}$ & 55.6 & 36.1 & 38.8 & 75.7 & 3.77 & $\mathcal{M}\,\mathcal{Q}\,\mathcal{C}\,\mathcal{W}$ & 37.0 & 39.1 & 25.5 & 32.6 & 22.14 \\
$\mathcal{Q}\,\mathcal{C}\,\mathcal{W}\,\mathcal{M}$ & 57.6 & 34.4 & 38.8 & 75.7 & 3.52 & $\mathcal{W}\,\mathcal{M}\,\mathcal{Q}\,\mathcal{C}$ & 35.3 & 41.1 & 24.9 & 33.4 & 22.07 \\
$\mathcal{Q}\,\mathcal{W}\,\mathcal{M}\,\mathcal{C}$ & 55.5 & 36.1 & 38.8 & 73.1 & 2.88 & $\mathcal{M}\,\mathcal{Q}\,\mathcal{W}\,\mathcal{C}$ & 37.0 & 41.1 & 25.5 & 30.8 & 21.91 \\
$\mathcal{Q}\,\mathcal{W}\,\mathcal{C}\,\mathcal{M}$ & 57.6 & 33.9 & 38.8 & 73.1 & 2.25 & $\mathcal{Q}\,\mathcal{W}\,\mathcal{M}\,\mathcal{C}$ & 33.5 & 41.1 & 26.1 & 33.3 & 21.87 \\
$\mathcal{C}\,\mathcal{W}\,\mathcal{Q}\,\mathcal{M}$ & 57.6 & 35.4 & 39.0 & 64.3 & 0.31 & $\mathcal{Q}\,\mathcal{M}\,\mathcal{C}\,\mathcal{W}$ & 35.2 & 39.1 & 26.1 & 32.6 & 21.39 \\
$\mathcal{W}\,\mathcal{C}\,\mathcal{Q}\,\mathcal{M}$ & 57.6 & 34.8 & 39.0 & 62.1 & -0.97 & $\mathcal{M}\,\mathcal{W}\,\mathcal{Q}\,\mathcal{C}$ & 37.0 & 41.1 & 24.9 & 30.9 & 21.35 \\
$\mathcal{W}\,\mathcal{Q}\,\mathcal{M}\,\mathcal{C}$ & 55.5 & 36.1 & 38.4 & 62.1 & -1.48 & $\mathcal{W}\,\mathcal{Q}\,\mathcal{M}\,\mathcal{C}$ & 33.5 & 41.1 & 25.4 & 33.4 & 21.31 \\
$\mathcal{W}\,\mathcal{Q}\,\mathcal{C}\,\mathcal{M}$ & 57.6 & 33.9 & 38.4 & 62.1 & -2.11 & $\mathcal{Q}\,\mathcal{M}\,\mathcal{W}\,\mathcal{C}$ & 35.2 & 41.1 & 26.1 & 30.8 & 21.16 \\
\bottomrule
\end{tabular}
\end{table}

\clearpage
\subsection{Detailed Hyper-Parameters}
We summarize the hyperparameters used in our experiments in Table \ref{tab:training-hparams}. These values were chosen through a combination of prior work, small-scale ablations, and practical compute considerations.
\begin{table*}[h]
    \centering
    \large
    \caption{Training Hyperparameters for All Training Settings. \textbf{ST} = Single-Task RL (e.g., \textbf{ST Math} = RL trained only on math). }

    \vskip 0.15in
    \label{tab:training-hparams}
    \resizebox{\textwidth}{!}{%
    \begin{tabular}{lccccccc}
        \toprule
        \textbf{Hyperparameter} & 
        \textbf{\shortstack{Curr.\\Learning}} & 
        \textbf{\shortstack{Joint\\Training}} & 
        \textbf{\shortstack{ST\\Coding}} & 
        \textbf{\shortstack{ST\\Math}} & 
        \textbf{\shortstack{ST\\QA}} & 
        \textbf{\shortstack{ST\\Writing}} & 
        \textbf{SFT} \\
        \midrule
        \multicolumn{8}{l}{\textit{Model Configuration}} \\
        \quad Max Prompt Length & 1024 & 1024 & 1024 & 1024 & 1024 & 1024 & - \\
        \quad Max Response Length & 3072 & 3072 & 3072 & 3072 & 3072 & 3072 & - \\
        \midrule
        \multicolumn{8}{l}{\textit{Training Settings}} \\
        \quad Train Batch Size & 256$\times$6 & 256$\times$6 & 256$\times$6 & 256$\times$6 & 256$\times$6 & 256$\times$6 & 128 \\
        \quad Learning Rate & 1e-6 & 1e-6 & 1e-6 & 1e-6 & 1e-6 & 1e-6 & 2.5e-6 \\
        \quad Learning Scheduler & Constant & Constant & Constant & Constant & Constant & Constant & Cosine \\
        \quad Optimizer & AdamW & AdamW & AdamW & AdamW & AdamW & AdamW & AdamW \\
        \quad Grad Clip & 1.0 & 1.0 & 1.0 & 1.0 & 1.0 & 1.0 & 1.0 \\
        \quad Max Epoch & 3 & 3 & 3 & 3 & 3 & 3 & 3 \\
        \midrule
        \multicolumn{8}{l}{\textit{RL Settings}} \\
        \quad KL Beta & 0.0 & 0.0 & 0.0 & 0.0 & 0.0 & 0.0 & - \\
        \quad Clip Ratio Low & 0.2 & 0.2 & 0.2 & 0.2 & 0.2 & 0.2 & - \\
        \quad Clip Ratio High & 0.2 & 0.2 & 0.2 & 0.2 & 0.2 & 0.2 & - \\

        \quad $N$ Rollouts & 16 & 16 & 16 & 16 & 16 & 16 & - \\
        \quad Rollout Temperature & 1.0 & 1.0 & 1.0 & 1.0 & 1.0 & 1.0 & - \\
        \quad Rollout Top-P & 1.0 & 1.0 & 1.0 & 1.0 & 1.0 & 1.0 & - \\
        \quad Rollout Top-K & 50 & 50 & 50 & 50 & 50 & 50 & - \\
        \midrule
        \multicolumn{8}{l}{\textit{LLM-as-a-Judge Settings}} \\
        \quad Model & gpt-4.1-mini & gpt-4.1-mini & - & - & - & gpt-4.1-mini & - \\
        \quad Temperature & 0.4 & 0.4 & - & - & - & 0.4 & - \\
        \bottomrule
    \end{tabular}
    }
\end{table*}

\end{document}